\documentclass{article}

\usepackage[numbers,sort&compress]{natbib}

\usepackage{pallas}

\usepackage[utf8]{inputenc}
\usepackage[T1]{fontenc}
\usepackage{hyperref}
\usepackage{url}
\usepackage{booktabs}
\usepackage{amsfonts}
\usepackage{amsmath}
\usepackage{amssymb}
\usepackage{nicefrac}
\usepackage{microtype}
\usepackage{xcolor}
\usepackage{graphicx}
\usepackage{algorithm}
\usepackage{algpseudocode}
\usepackage{enumitem}
\usepackage{xspace}
\usepackage{listings}
\usepackage{wrapfig}
\usepackage{afterpage}
\usepackage{placeins}

\definecolor{task_hover}{RGB}{141,168,200}    
\definecolor{task_codeio}{RGB}{125,191,158}   
\definecolor{task_polaris}{RGB}{207,178,99}   
\definecolor{task_physics}{RGB}{232,144,124}  
\definecolor{promptbreak}{RGB}{153,153,153}  
\colorlet{promptbg}{task_hover}
\colorlet{promptframe}{task_hover}

\newcommand{\promptbox}[3]{%
  \colorlet{promptbg}{#3}%
  \colorlet{promptframe}{#3}%
  \par\noindent
  \begingroup
  \setlength{\fboxsep}{6pt}%
  \colorbox{promptframe!85!black}{%
    \parbox{\dimexpr\linewidth-2\fboxsep\relax}{%
      \color{white}\bfseries\small
      \rule[-4pt]{0pt}{14pt}#1\strut}%
  }\par\nointerlineskip
  \endgroup
  \lstinputlisting[style=gepaprompt]{#2}%
  \medskip%
}

\lstdefinestyle{gepaprompt}{
  basicstyle=\ttfamily\fontsize{8pt}{9.5pt}\selectfont,
  backgroundcolor=\color{promptbg!60!white},
  frame=single,
  rulecolor=\color{promptframe!85!black},
  framerule=0.3mm,
  breaklines=true,
  breakindent=0pt,
  columns=fullflexible,
  keepspaces=true,
  showstringspaces=false,
  xleftmargin=0pt,
  xrightmargin=0pt,
  framesep=0pt,
  aboveskip=0pt,
  belowskip=0pt,
  literate={–}{{-}}1 {—}{{---}}1 {’}{{'}}1 {‘}{{'}}1
           {“}{{"}}1 {”}{{"}}1 {…}{{...}}1
           {↔}{{<->}}1 {→}{{$\to$}}1 {⇒}{{$\Rightarrow$}}1
           {Δ}{{$\Delta$}}1 {π}{{$\pi$}}1
           {≥}{{$\geq$}}1 {≤}{{$\leq$}}1
           {×}{{$\times$}}1 {±}{{$\pm$}}1 {°}{{$^\circ$}}1
           {²}{{$^2$}}1 {³}{{$^3$}}1
           {ℕ}{{$\mathbb{N}$}}1 {ℝ}{{$\mathbb{R}$}}1,
}

\definecolor{linkblue}{HTML}{0060E0}
\hypersetup{urlcolor=linkblue,pdftitle={Learning, Fast and Slow: Towards LLMs That Adapt Continually}}

\providecommand{\methodname}{\textsc{FST}\xspace}

\begin{document}
\addtocontents{toc}{\protect\setcounter{tocdepth}{-1}}


\begin{tcolorbox}[abstractbox, width=\textwidth]
  \centering

  {\LARGE\titleFont \textcolor{titleblue}{Learning, Fast and Slow: Towards LLMs\\ That Adapt Continually}\par}
  \vspace{0.5em}

{\normalsize\authorFont
  \textbf{Rishabh Tiwari}$^{*\,1, 4}$\hspace{0.2em}
  \textbf{Kusha Sareen}$^{*\,2}$\hspace{0.2em}
  \textbf{Lakshya A Agrawal}$^{*\,1}$\\[0.2em]
  \textbf{Joseph E. Gonzalez}$^{\,1}$\hspace{0.2em}
  \textbf{Matei Zaharia}$^{\,1}$\hspace{0.2em}
  \textbf{Kurt Keutzer}$^{\,1}$\hspace{0.2em}
  \textbf{Inderjit S Dhillon}$^{\,3}$\\[0.2em]
  \textbf{Rishabh Agarwal}$^{\dagger\,2, 5}$\hspace{0.2em}
  \textbf{Devvrit Khatri}$^{\dagger\,3, 6}$
}\\[0.8em]
{\normalsize\authorFont $^{1}${UC Berkeley}\hspace{1.5em}$^{2}${Mila}\hspace{1.5em}$^{3}${UT Austin}\hspace{1.5em}$^{4}${Eragon}\hspace{1.5em}$^{5}${Periodic Labs}\hspace{1.5em}$^{6}${Mirendil}}\\[0.3em]
  {\normalsize\authorFont\hypersetup{urlcolor=titleblue} \href{https://rishabhtiwari.ai/projects/fst/video.mp4}{Video} \hspace{0.5em}|\hspace{0.5em} \href{https://gepa-ai.github.io/gepa/blog/2026/05/11/learning-fast-and-slow/}{Blog} \hspace{0.5em}|\hspace{0.5em} \href{https://rishabhtiwari.ai/projects/fst/code/}{Code}}\\[0.1em]

  \vspace{0.1em}
  \begin{justify}
  \authorFont
  \setstretch{1.1}
  \setlength{\parindent}{0pt}
Large language models (LLMs) are trained for downstream tasks by updating their parameters (e.g., via RL). However, updating model parameters forces them to absorb task-specific information, which can result in catastrophic forgetting and loss of plasticity.
In contrast, in-context learning with fixed LLM parameters can cheaply and rapidly \emph{adapt} to task-specific requirements (e.g., prompt optimization), but cannot by itself typically match the performance gains available through updating LLM parameters.
Inspired by how human learning works at multiple time scales (e.g., System 1 vs 2), we propose a fast-slow learning framework for LLMs, instantiated with model parameters as ``slow'' weights and optimized context as ``fast'' weights. These fast ``weights'' can learn from textual feedback to absorb the task-specific information, while allowing slow weights to stay closer to the base model and persist general reasoning behaviors.
\textbf{Fast-Slow Training} (\methodname{}) is up to $3\times$ more sample-efficient than only slow learning (RL) across reasoning tasks, while consistently reaching a higher performance asymptote. Moreover, FST-trained models remain closer to the base LLM (up to $70\%$ less KL divergence), resulting in less catastrophic forgetting than RL-training. 
This reduced drift also preserves plasticity: after training on one task, FST trained models adapt more effectively to a subsequent task.
In continual learning scenarios, where task domains change on the fly, FST continues to acquire each new task while RL stalls.

   \end{justify}

  \vspace{0.3em}
  {\centering\small\authorFont \textbf{Correspondence}: \texttt{\href{mailto:rishabhtiwari@berkeley.edu,lakshyaaagrawal@berkeley.edu}{\{rishabhtiwari, lakshyaaagrawal\}@berkeley.edu}, \href{mailto:kusha.sareen@mila.quebec}{kusha.sareen@mila.quebec}, \href{mailto:devvrit.03@gmail.com,rishabhagarwal.467@gmail.com}{\{devvrit.03, rishabhagarwal.467\}@gmail.com}\\[0.5em]$^{*}$Equal contribution\hspace{0.1em} $^{\dagger}$Equal advising.}\par}
\end{tcolorbox}
\vspace{1.0em}
\vspace{-1.5mm}
\section{Introduction}
\vspace{-1.5mm}
\begin{figure}[t]
    \centering
    \includegraphics[width=\linewidth]{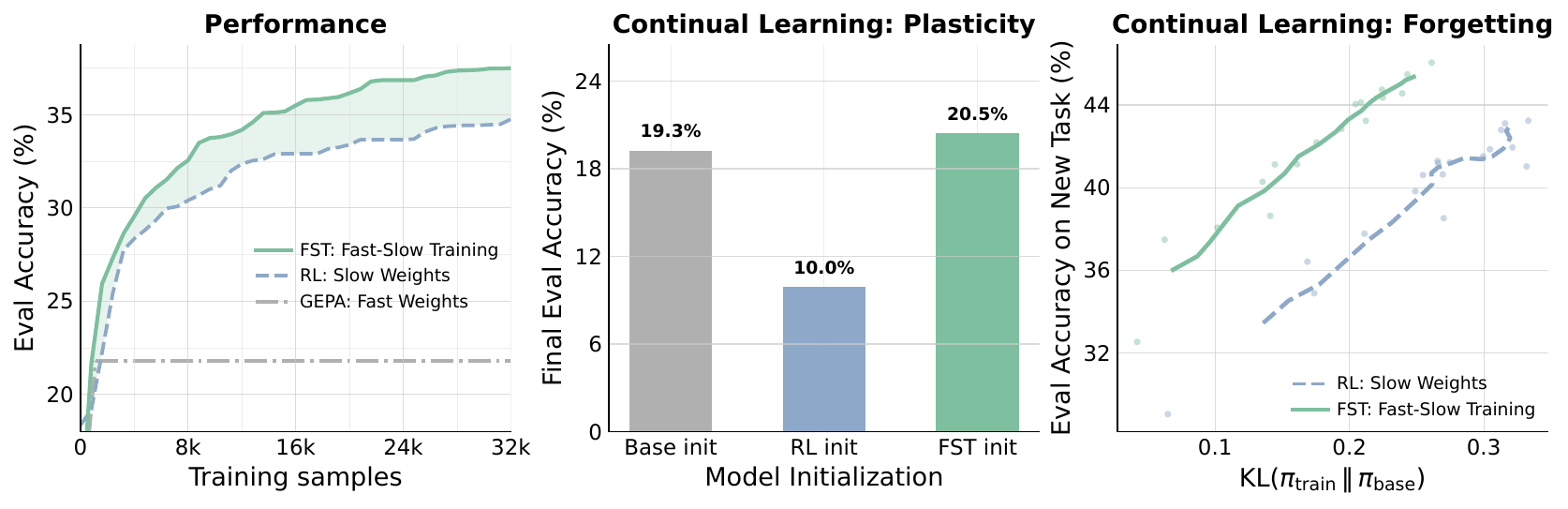}
    \vspace{-4.5mm}
    \caption{\textbf{Fast-Slow Training (\methodname{}) learns faster, stays adaptable, and forgets less.} Comparison of \methodname{} (slow-weight RL interleaved with fast-weight prompt optimization), RL alone (slow weights only), and GEPA alone (fast weights only), averaged over \texttt{CodeIO}, \texttt{Math (Polaris)}, and \texttt{HoVer-hard}.
    \textbf{Left:} Evaluation accuracy on the trained task as training samples accumulate. \methodname{} reaches RL's peak with substantially fewer samples and converges to a higher ceiling than either RL or GEPA alone.
    \textbf{Middle:} Plasticity, the model's remaining ability to learn a new skill. After training on a first task, we continue with a fresh round of RL on a second task and report the final accuracy from each initialization. The RL-trained checkpoint barely learns the new task, while the \methodname{}-trained checkpoint roughly matches the base model.
    \textbf{Right:} How far each training run drifts from the base model, measured by KL$(\pi_{\mathrm{train}}\,\|\,\pi_{\mathrm{base}})$. Smaller drift correlates with less forgetting of prior abilities; at matched accuracy, \methodname{} sits well to the left of RL.}
    \label{fig:hero}
\end{figure}

\afterpage{\afterpage{%
\begin{figure}[t]
    \centering
    \includegraphics[width=\linewidth]{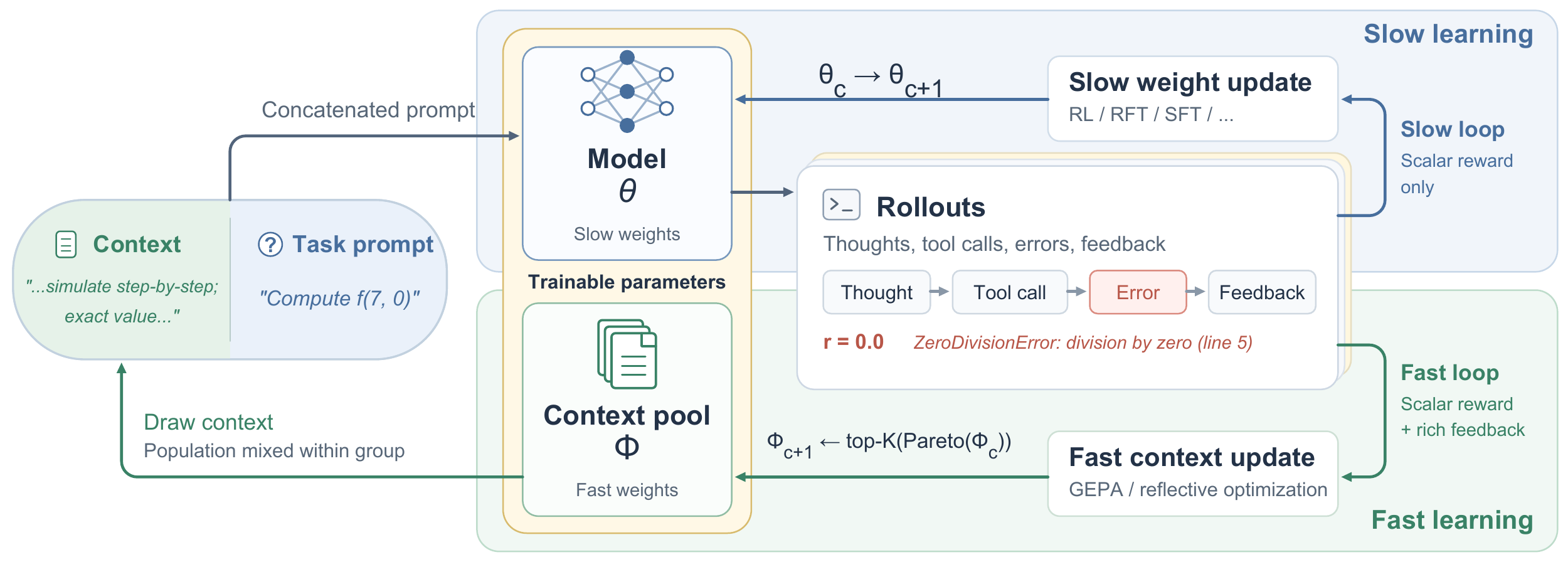}
    \vspace{-4.5mm}
    \caption{\textbf{Slow weights and fast weights co-evolve through interleaved updates.} The slow loop (top) updates $\theta$ from the scalar reward alone ($\theta_c \to \theta_{c+1}$). The fast loop (bottom) updates $\Phi$ via reflective optimization, additionally consuming the rollout's full text including thoughts, tool calls, errors, and rich feedback ($\Phi_c \to \Phi_{c+1}$). Maintaining $\Phi$ as a Pareto-frontier population (rather than a single best prompt) preserves diversity: different contexts specialize to different problem slices exposing the slow update rule to rich conditioning during training.
    }
    \label{fig:method}
\end{figure}
}}

Large language models (LLMs) are commonly adapted through supervised finetuning (SFT) or reinforcement learning (RL), both of which modify the model parameters, to specialized domains such as math and coding~\citep{deepseek_r1,tulu3,deepseekmath,deepseekcoder,instructgpt}. However, treating model parameter updates as the sole mechanism of adaptation creates a fundamental bottleneck: every improvement, whether it be a reusable reasoning skill, a task-specific heuristic or a transient lesson from recent rollouts, must be written into the same persistent set of model weights.
Since the entire policy is parameterized by these weights, an update that improves in-domain reward simultaneously
moves the model away from its base behavior~\citep{gao2022scalinglawsrewardmodel,mitigatingalignmenttaxrlhf}, reducing entropy~\citep{entropymechanism,li2026choicedivergenceneglectedkey}, hurting out-of-distribution generalization~\citep{instructgpt,scalinglawsforgettingfinetuning,empiricalstudycatastrophicforgetting,beyondreasoninggains}, and degrading its ability to adapt to future tasks, known as plasticity loss \citep{Frati_2024,understandingpreventingcapacityloss,Dohare2024,mitigatingplasticitylosscontinual}.

LLM systems also possess another powerful adaptation mechanism: prompts, instructions, and contextual information~\citep{languagemodelsfewshotlearners,cot}. Unlike model parameters, these textual components can be modified cheaply, frequently, and per task. Prompt optimization methods
demonstrate that substantial behavioral improvements can be obtained by improving the textual context under which the model operates~\citep{ape,opro,dspy,miprov2,gepa}.

We introduce \textbf{Fast-Slow Training (FST)}, a general framework that interleaves optimization of model parameters (\emph{slow weights}) and adaptable textual components (\emph{fast weights}) as shown in Figure~\ref{fig:method}. Fast weights can include prompts, context, tool definitions, or the text and code defining an agent harness, including its control flow. Slow weights can be updated through RL, SFT, or preference optimization. For the scope of this paper, we focus on an RL + GEPA instantiation, rather than all these possibilities.


The distinction between fast and slow-moving weights has a long history in neural networks~\citep{hinton1987using,schmidhuber1992learning,ba2016usingfastweightsattend}. A common motivation across this line of work is to separate different forms of learning: temporary, task-specific, or lightweight adaptations can be captured by fast weights, while more persistent and broadly useful behaviors are consolidated into slow weights. Motivated by this perspective, we adopt a fast-slow weight view of LLM post-training. In our setting, model parameters naturally serve as slow weights: they are expensive to update, persist across tasks, and encode long-lived behavior. In contrast, optimized textual context, such as prompts, instructions, and task-specific feedback, acts as a form of fast weight: it can be changed cheaply and frequently, influences model behavior immediately, and captures task-level adaptation without permanently modifying the model.


This division of labor has several consequences, which we evaluate in RL settings spanning math, code, general reasoning and non-verifiable rubric-reward tasks.
\begin{enumerate}[leftmargin=*, itemsep=0pt,topsep=1pt]
    \item \textbf{Fast textual adaptation improves data efficiency.} Fast-slow training matches RL reward with up to $3\times$ fewer rollouts and consistently reaches a higher performance ceiling (Section~\ref{sec:de} Advantages 1 and 2).
    \item \textbf{Fast-slow training induces smaller slow-weight displacement.} With the textual channel carrying part of the adaptation, the model parameters need not move as far from the base policy. At matched reward, our models have up to 70\% lower KL to the base policy than RL-only baselines. (Section~\ref{sec:displacement} Advantage 3)
    \item \textbf{Fast-slow training preserves plasticity.}
    After training on one task, FST models adapt effectively to a second task, while RL-only models collapse to near $0\%$, suggesting greater capacity for future learning with FST (Section~\ref{sec:plasticity} Advantage 4).
    \item \textbf{Fast-slow training enables continual learning.} FST adapts more quickly to changing objectives when tasks change during training (Section~\ref{sec:cl} Advantage 5).
\end{enumerate}

    Our results support jointly training fast textual and slow parametric weights, rather than learning model parameters and then tuning context, to improve task performance while preserving generality and plasticity.

\section{Preliminaries}
\label{sec:prelim}


\paragraph{Fast and slow weights: a general framework}

We model \emph{slow weights} (model parameters) as $\theta$ and \emph{fast weights} (adaptable text or code) as $\phi\in\Sigma^\ast$. Optimizing $\phi$ can mean prompt or context optimization, tool optimization, or full harness optimization when the harness is represented as text or code. Given a query $x$, the system produces a response by sampling
\begin{equation}
    y \sim \pi_\theta(\cdot \mid x, \phi),
    \label{eq:policy}
\end{equation}


where $\pi_\theta(y \mid x, \phi)$ denotes the response distribution of the system defined by $\theta$ and $\phi$ on query $x$. For a task distribution $\mathcal{D}$ and
reward $r$, the natural joint objective is
\begin{equation}
    \max_{\theta,\, \phi}\;\; J(\theta, \phi) \;=\; \mathbb{E}_{x \sim \mathcal{D},\; y \sim \pi_\theta(\cdot \mid x, \phi)}\!\big[\,r(x, y)\,\big].
    \label{eq:joint-base}
\end{equation}
The framework is optimizer-agnostic. Slow weights can use SFT, preference optimization~\cite{dpo}, or policy-gradient methods such as PPO~\cite{ppo} and GRPO~\cite{deepseekmath}, often with verifiable rewards~\cite{tulu3}. Fast weights can use prompt/harness optimizers such as APE~\cite{ape}, OPRO~\cite{opro}, DSPy/MIPROv2~\cite{dspy,miprov2}, Meta-Harness~\cite{lee2026meta}, and GEPA~\cite{gepa}. We instantiate FST with RL with verifiable rewards (RLVR) and GEPA.

\paragraph{Slow weights: RL with verifiable rewards}
\label{sec:prelim:slow}

We follow ScaleRL~\cite{scalerl}, using verifier rewards $r(x,y)\in[0,1]$~\cite{tulu3}, such as rule-based correctness for math, code, and science. We additionally explore non-verifiable domains with rubric rewards in Appendix \ref{app:nonverifiable}. For each query $x$, the current policy generates $G$ rollouts $\{y_i\}_{i=1}^G$ under $(\theta,\phi)$, with group-relative advantages~\cite{deepseekmath}
\begin{equation}
    A_i \;=\; \frac{r(x, y_i) \;-\; \bar r_g}{\sigma_g \;+\; \varepsilon}, \qquad \bar r_g \;=\; \tfrac{1}{G}\!\sum_{j=1}^{G} r(x, y_j), \qquad \sigma_g^2 \;=\; \tfrac{1}{G}\!\sum_{j=1}^{G}\!\big(r(x, y_j) - \bar r_g\big)^2, \label{eq:adv}
\end{equation}
and normalized at the batch level. The policy is updated using the truncated importance-sampling REINFORCE objective \textsc{cispo}~\cite{minimaxm1,scalerl},
\begin{equation}
    \mathcal{L}_{\textsc{cispo}}(\theta) \;=\; -\,\mathbb{E}\!\left[\, \mathrm{sg}\!\big(\min(\rho_t,\,\tau)\big)\,\cdot\, A \,\cdot\, \log \pi_\theta(y_t \mid x, \phi, y_{<t})\,\right],
    \label{eq:cispo}
\end{equation}
where $\rho_t = \pi_\theta(y_t \mid x, \phi, y_{<t}) / \pi_{\theta_{\text{old}}}(y_t \mid x, \phi, y_{<t})$ is the per-token importance ratio between the current and rollout policies, $\tau$ the truncation threshold, and $\mathrm{sg}(\cdot)$ the stop-gradient operator. Loss is aggregated at the prompt level. Conventional RLVR updates only $\theta$, keeping $\phi$ fixed to a generic system prompt.

\paragraph{Fast weights: reflective prompt evolution.}
We optimize the fast weights $\phi$ using GEPA~\cite{gepa}, a reflective evolutionary procedure over textual prompts $\phi \in \Sigma^\ast$. For a fixed policy $\pi_\theta$, the fitness of a prompt on instance $x$ is its expected reward,

\begin{equation}
    s(\phi; x) \;=\; 
    \mathbb{E}_{y \sim \pi_\theta(\cdot \mid x, \phi)}
    \!\left[ r(x,y) \right].
    \label{eq:fitness}
\end{equation}
GEPA maintains a population of prompts, uses rollouts to elicit natural-language critiques from a frozen reflection LM, and proposes textual mutations that improve performance on an anchor set from $\mathcal{D}$. Rather than returning a single prompt, GEPA retains a Pareto frontier of complementary prompts and returns the top-$m$ candidates, which we use as fast weights. We defer the details of parent selection, mutation, pruning, and prompt examples to Appendix~\ref{appendix:gepa}.

\section{Fast-Slow Training (FST)}
\label{sec:method}

We now describe the RL + GEPA instantiation of \methodname{}. It updates slow weights $\theta$ through RL and a population of fast weights $\Phi$ through GEPA. This instantiation maintains $K$ textual prompts,
$\Phi=\{\phi^{(1)},\ldots,\phi^{(K)}\}$, and optimizes
\begin{equation}
    \max_{\theta,\,\Phi}\;\;
    J(\theta,\Phi)
    =
    \mathbb{E}_{x\sim\mathcal{D},\;\phi\sim U(\Phi),\;y\sim\pi_\theta(\cdot\mid x,\phi)}
    \!\left[r(x,y)\right],
    \label{eq:joint-pop}
\end{equation}
where $U(\Phi)$ is uniform over the prompt population. We keep a population rather than one best prompt (examples of evolved prompts are in Appendix~\ref{app:gepa-prompts}). GEPA returns a Pareto frontier of complementary prompts: different prompts excel on different subsets of $\mathcal{D}$. Sampling across this frontier during RL gives the policy access to multiple conditioning behaviors and lets group-relative advantages compare both prompt-induced and sampling-induced variation on the same problem.

Training proceeds in cycles of $T$ slow-weight updates. At the start of cycle $c$, we pre-fetch the next $T$ RL batches and denote their union by the lookahead batch $\mathcal{L}_c$. We run GEPA with the current policy $\pi_{\theta_c}$ as the rollout model, a frozen reflection LM $\pi_{\mathrm{ref}}$ as the proposer, $\mathcal{L}_c$ or a fixed-size subset as the anchor set, and the previous population $\Phi_c$ as the seed. GEPA returns the top-$K$ candidates from its Pareto frontier, yielding the fast weights $\Phi_{c+1}$.

For the next $T$ steps, we update $\theta$ on minibatches from $\mathcal{L}_c$ while holding $\Phi_{c+1}$ fixed. For each problem $p$, we form a rollout group of size $G$ by sampling each prompt $\phi^{(k)}\in\Phi_{c+1}$ exactly $G/K$ times. That is, in each group, $G/K$ rollouts receive the same prompts and we have $K$ such mini-groups. Cumulatively, they are treated as one group for $p$; rewards are normalized by the per-problem statistics $(\bar r_g,\sigma_g)$ as in eq~\ref{eq:adv}, mixing prompt and sampling variation within the same advantage computation. We then apply the \textsc{cispo} update in Eq.~\eqref{eq:cispo}. After $T$ updates, the procedure repeats with a new GEPA phase under the updated policy. Pseudocode of \methodname is given in Appendix~\ref{app:algorithm}.

\section{Advantages of Fast-Slow Training} \label{sec:benefits}

The textual fast weights $\Phi$ carry part of the task-level information that RL would otherwise force into $\theta$, so the slow weights move less to reach the same reward. The downstream signature of this division of labor is consistent across our settings: training reaches matched reward more quickly, $\theta$ drifts less from the base policy at convergence, the model retains greater plasticity to adapt to subsequent tasks, and our method shows higher continual learning capability. We show each in the following sections.

\paragraph{Advantage 1: Fast-Slow Training Improves Data Efficiency}\label{sec:de}

\begin{figure}[t]
    \centering
    \includegraphics[width=\linewidth]{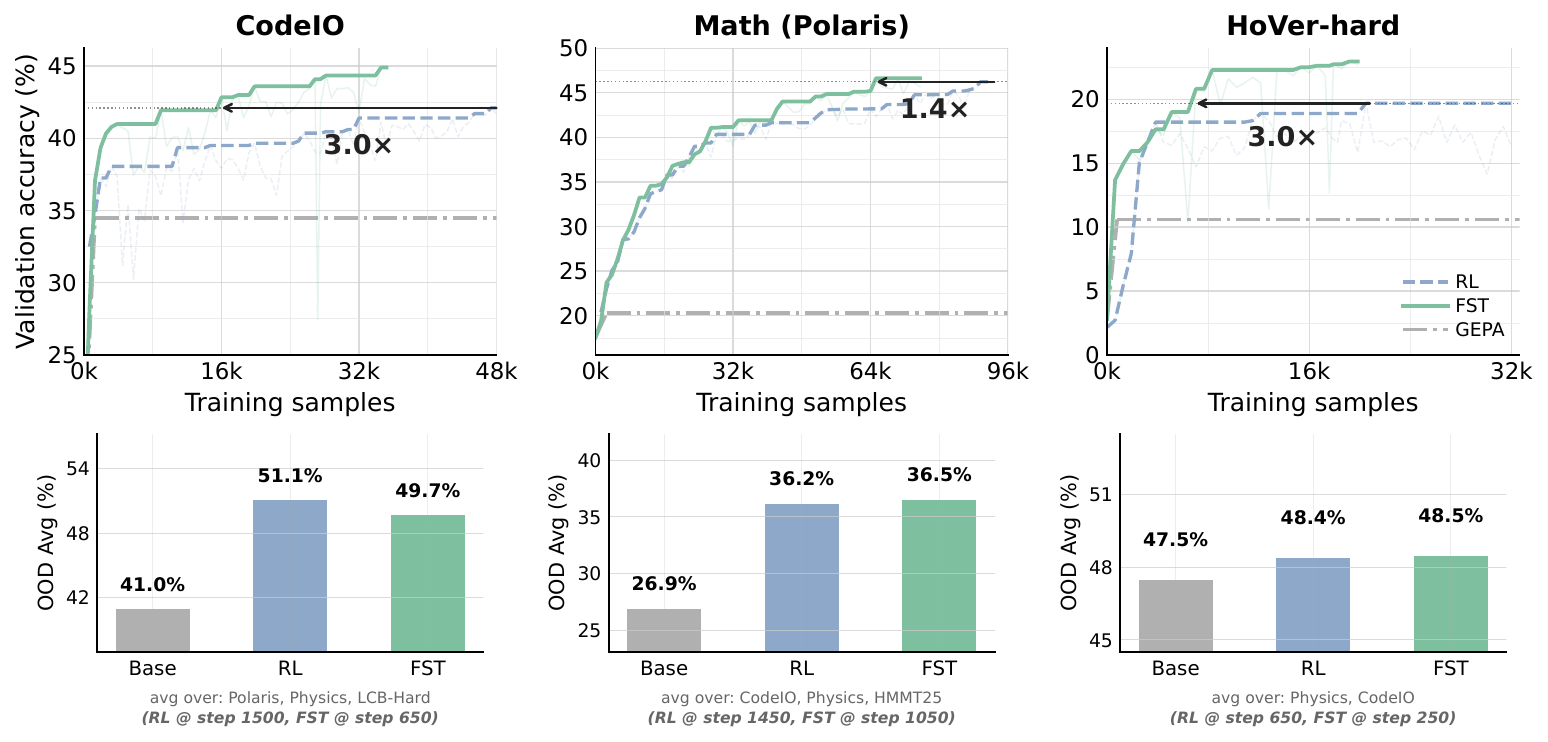}
    \vspace{-4.5mm}
    \caption{\textbf{Data efficiency across three training families.} \textbf{Top row}: matched-step validation accuracy (running max, mean@4); dash-dot GEPA-only reference rises from the step-0 base accuracy to the prompt-only ceiling within GEPA's inference budget. \methodname{} reaches RL's running peak in substantially fewer training steps ($3.0\times$ on \texttt{CodeIO}, $1.4\times$ on \texttt{Math} (Polaris), $3.0\times$ on \texttt{HoVer-hard}). \textbf{Bottom row}: out-of-distribution accuracy averaged across cross-domain (and easy$\to$hard, where available) benchmarks for each family, evaluated with no GEPA prompt. \methodname{} matches RL on OOD averages while reaching the in-distribution peak with substantially fewer  steps.}
    \label{fig:data-efficiency}
\end{figure}

We evaluate \methodname{} on three training families: code-output prediction (\texttt{CodeIO})~\citep{codeio,reasoning_gym}, math (\texttt{Polaris})~\citep{polaris}, and multi-hop fact verification (\texttt{HoVer-hard})~\citep{hover}. All experiments use \texttt{Qwen3-8B}~\citep{qwen3}, except for the Math run, where we first SFT \texttt{Qwen3-8B-Base} on Nemotron data~\citep{nemotronmath} because \texttt{Qwen3-8B} is already saturated on math benchmarks (Qwen3-32B, Qwen3-4B, and OLMo-3-7B results are in Appendices~\ref{app:model-size} and~\ref{app:nonverifiable}). \methodname{} uses cycle length $T{=}6$ and $K \in \{4, 8\}$ candidate prompts per cycle. Training-time performance is measured on a held-out in-distribution validation set. RL is trained until step $1500$ or in-distribution saturation (whichever comes first); \methodname{} is trained at least until it matches RL's running peak.
Full hyperparameters and dataset details are deferred to Appendix~\ref{app:hyperparams} and cost to Appendix~\ref{app:rollout-reuse}.

The matched-step training curves (Figure~\ref{fig:data-efficiency} Top) show that \methodname{} reaches RL's running peak in substantially fewer steps: $\mathbf{3.0\times}$ fewer on \texttt{CodeIO}
, $\mathbf{1.4\times}$ on \texttt{Math} 
, and $\mathbf{3.0\times}$ on \texttt{HoVer-hard}. Continuing past the crossover, \methodname{}'s running peak also exceeds RL's on all three tasks 
.

To check that the in-distribution data efficiency does not come at the cost of out-of-distribution behavior, for each training task we compare Base, RL's final checkpoint, and \methodname{}'s validation matched-performance checkpoint on a family of cross-domain and easy-to-hard generalization datasets, with no GEPA prompt at inference (Figure~\ref{fig:data-efficiency} bottom). Across all three training families, \methodname{} reaches RL's in-distribution peak with $1.4\times$ to $3.0\times$ fewer training samples while retaining comparable OOD performance. 
Thus, \methodname{}'s faster in-distribution learning is accompanied by preserved cross-domain and easy-to-hard generalization.

\paragraph{Advantage 2: Fast-Slow Training Raises the Performance Asymptote}\label{sec:asymptote}

\begin{figure}[t]
    \centering
    \includegraphics[width=\linewidth]{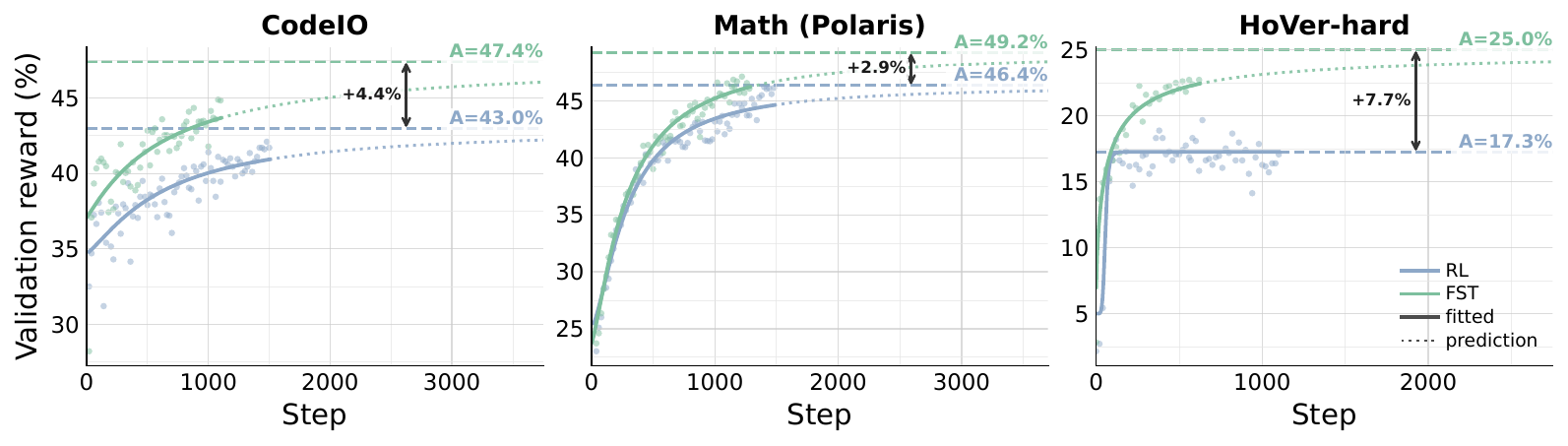}
    \vspace{-4.5mm}
    \caption{\textbf{Performance asymptote} on \texttt{CodeIO}, \texttt{Math} (Polaris), and \texttt{HoVer-hard}. For each run we fit a 4-parameter sigmoid $R-R_0 = \frac{A - R_0}{1 + (C_{mid}/C)^B}$ to the validation-accuracy trajectory and annotate the upper asymptote $A$. \methodname{}'s asymptote (green) is higher than RL's (blue) on all three tasks. Solid curves cover the fit window; dotted curves are extrapolation past the last training step.}
    \label{fig:performance-asymptote}
\end{figure}

Following \citet{scalerl}, we compare RL and \methodname{} by the saturation level of their validation-accuracy curves rather than at any single training step. Unlike final-step or matched-step accuracy, which depends on where each run was stopped, the asymptote of a fitted curve reads off the level the run is converging to. For each (task, method) we fit a sigmoid curve 
\begin{equation}
    \Delta R = \frac{A - R_0}{1 + (C_{mid}/C)^B}
\end{equation}
to the validation-accuracy trajectory, where $A$ is the upper asymptote, $B$ a scaling exponent, $C_{mid}$ the midpoint of the performance, and $R_0$ is the initial reward at step 0.

Across all three tasks (Figure~\ref{fig:performance-asymptote}), \methodname{}'s fitted asymptote exceeds RL's: $A{=}47.4\%$ vs $43.0\%$ on \texttt{CodeIO} ($+4.4$pp), $49.2\%$ vs $46.4\%$ on \texttt{Math} (Polaris) ($+2.9$pp), and $25.0\%$ vs $17.3\%$ on \texttt{HoVer-hard} ($+7.7$pp). Pushing part of the task adaptation into the textual fast-weight channel $\Phi$ in addition to the slow weights $\theta$ helps the overall method converge to a higher accuracy ceiling than RL alone reaches; the same fit at Qwen3-32B is in Appendix~\ref{app:model-size}.


\paragraph{Advantage 3: Fast-Slow Training Remains Close to the Base Model}\label{sec:displacement}

\begin{figure}[t]
    \centering
    \includegraphics[width=\linewidth]{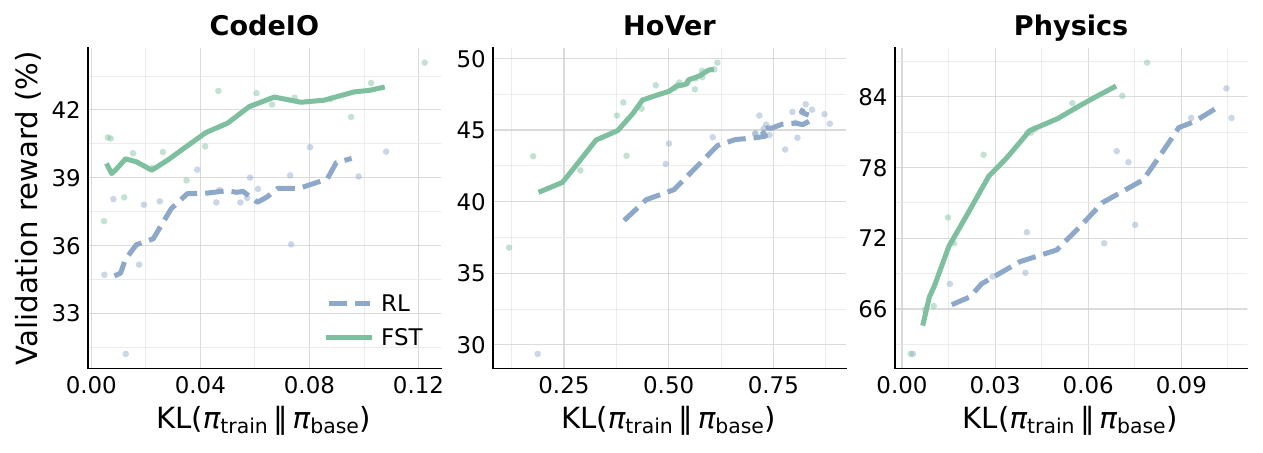}
    \vspace{-4.5mm}
    \caption{\textbf{Validation reward versus $\mathrm{KL}(\pi_{\text{train}}\,\|\,\pi_{\text{base}})$ trajectories} on \texttt{CodeIO}, \texttt{HoVer}, and \texttt{Physics}. Translucent markers are per-checkpoint measurements; the line is a rolling-mean smoothing along training step. At matched reward, \methodname{} (green) sits to the left of RL (blue) on every task, reaching the same reward at a significantly lower KL from the base policy. Full figure in Appendix~\ref{app:polaris-kl}.}
    \label{fig:kl-vs-acc}
\end{figure}

The KL divergence $\mathrm{KL}(\pi_{\text{train}}\,\|\,\pi_{\text{base}})$ between the post-trained policy and the base measures how far the slow weights have moved away from their base configuration; larger displacement is associated with 
reduced entropy,
weaker OOD generalization, and lower plasticity for future tasks~\citep{Frati_2024,understandingpreventingcapacityloss,Dohare2024,mitigatingplasticitylosscontinual}. We track this directly - at each training checkpoint we compute token-level KL from the base on the held-out validation prompts and plot it against the same checkpoint's validation accuracy, for both \methodname{} and RL across \texttt{Physics}, \texttt{Math} (Polaris), \texttt{HoVer}, and \texttt{CodeIO}.

Across all three tasks (Figure~\ref{fig:kl-vs-acc}), and on a rubric-scored task without a verifier (Appendix~\ref{app:nonverifiable}), \methodname{} achieves higher performance at 
lower KL than RL.
\citet{shenfeld2025rlrazor} recently showed that on-policy RL is already biased toward KL-minimal solutions on a new task, and that the size of this shift correlates with how much prior knowledge is forgotten. Even relative to this strong baseline, \methodname{} shifts the accuracy/KL frontier further left.
We next show that models trained with \methodname{} preserve plasticity and learn better continually due to reduced displacement and continually adapted fast weights (Appendix~\ref{app:cl-baselines}).


\paragraph{Advantage 4: Fast-Slow Training Preserves Plasticity}\label{sec:plasticity}

\begin{figure}[t]
    \centering
    \includegraphics[width=0.9\linewidth]{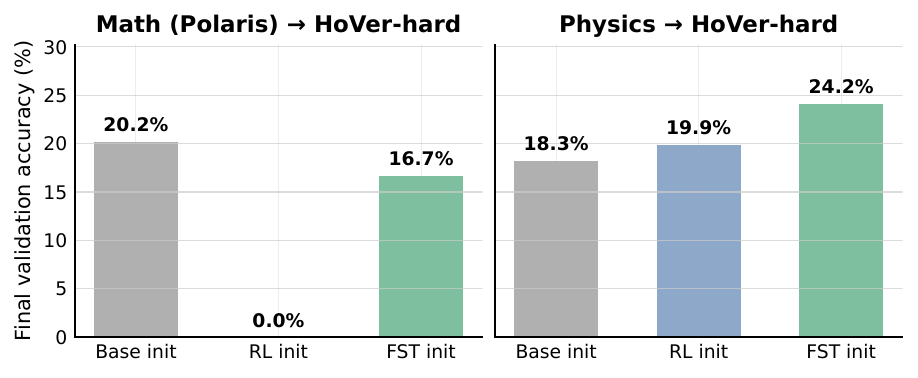}
    \vspace{-4.5mm}
    \caption{\textbf{Plasticity probe}: starting from a Math (left) or Physics (right) checkpoint trained with either RL or \methodname{}, we run a \emph{fresh} RL pass on \texttt{HoVer-hard} and plot HoVer validation accuracy over 400 steps. Base init (dotted) is the no-prior-training reference. \methodname{}-init (green) preserves more capacity for the new task than RL-init (blue) on both arms; on the Math arm, prior RL collapses \texttt{HoVer-hard} learnability to near-zero.}
    \label{fig:plasticity}
\end{figure}


Continued post-training has been observed to hamper a model's ability to learn future tasks, a phenomenon commonly called \emph{plasticity loss}~\citep{Dohare2024,understandingpreventingcapacityloss,Frati_2024,mitigatingplasticitylosscontinual}: the slow weights become specialized to the trained task and lose responsiveness to gradient signals from new ones.
We probe this directly in two phases.
\emph{Phase~1} trains a base model on task $X$ using either standard RL or \methodname{}.
\emph{Phase~2} takes the Phase-1 checkpoint, at the same training step for both arms, as initialization and runs standard RL on a different task $Y$.
Throughout Phase~2 we track validation accuracy on $Y$.
As a no-prior-training reference, we also run Phase~2 starting from the base model.
We test $\texttt{Math} \to \texttt{HoVer-hard}$ and $\texttt{Physics} \to \texttt{HoVer-hard}$.

Figure~\ref{fig:plasticity} shows that in Phase-2, \methodname{}-init outperforms RL-init through the 400-step probe in both settings. The contrast is sharpest in \texttt{Math}~$\to$~\texttt{HoVer-hard}: prior RL collapses \texttt{HoVer-hard} learnability to near-zero, the RL-init curve drops to $\sim$$0\%$ within 40 steps and stays flat for the rest of the run.
In contrast, \methodname{}-init reaches 
performance
close to the 
base-init reference.
On \texttt{Physics}~$\to$~\texttt{HoVer-hard}, \methodname{}-init finishes at $24.2\%$ and is still climbing, versus RL-init's $19.9\%$ at step 400 
.
This indicates that, unlike RL, \methodname{} does not over-specialize the slow weights to task $X$: the resulting checkpoint retains capacity to learn a new task $Y$, exhibiting higher plasticity.



\paragraph{Advantage 5: Fast-Slow Training Improves Continual Learning}\label{sec:cl}

\begin{figure}[t]
    \centering
    \includegraphics[width=\linewidth]{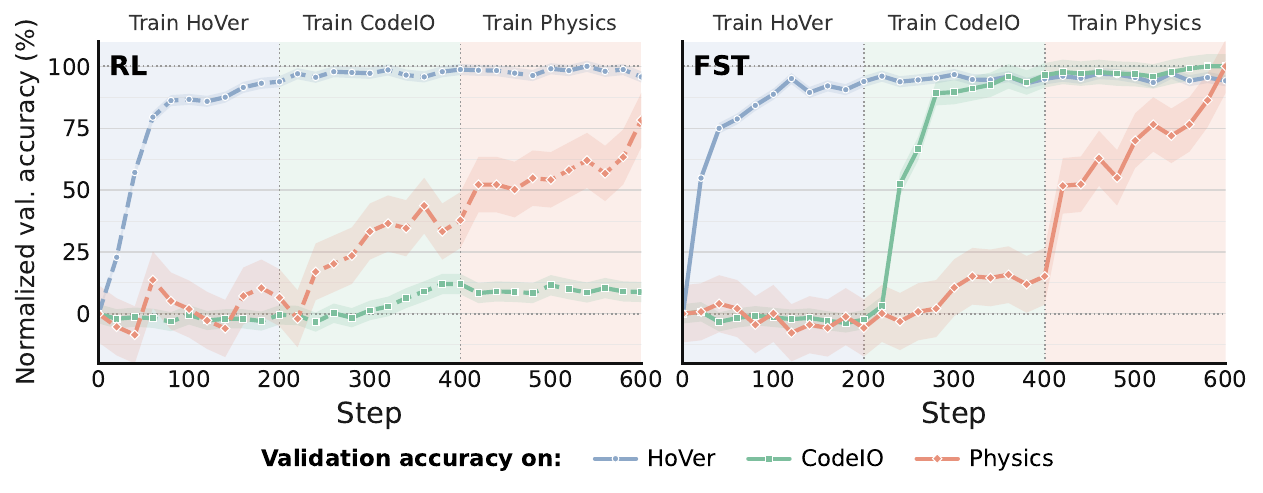}
    \vspace{-5mm}
    \caption{\textbf{Continual learning across \texttt{HoVer} $\to$ \texttt{CodeIO} $\to$ \texttt{Physics}}: a single uninterrupted training run that switches task every 200 steps. The y-axis is per-task validation accuracy, normalized so that $0$ is the base model's accuracy on the task and $100$ is the peak reached on that task by either method; bands are $\pm 1$ standard error over the validation rollouts. \methodname{} (solid) reaches near-peak on every stage; RL (dashed) acquires \texttt{HoVer} but completely stalls on \texttt{CodeIO} and only partially recovers on \texttt{Physics}.}
    \label{fig:cl}
\end{figure}

A continual learning algorithm must keep absorbing new tasks as training proceeds, without losing the capacity to absorb later ones~\citep{Dohare2024,mitigatingplasticitylosscontinual,understandingpreventingcapacityloss}.
To test this we run a single uninterrupted training pass over three tasks, sequentially swapping the task every 200 steps - first 200 steps with \texttt{HoVer} (multi-hop fact verification), then \texttt{CodeIO} (code-output prediction), and finally \texttt{Physics} (multiple-choice from \texttt{sciknoweval}).
In this setting, the same live training trajectory must absorb three task changes back-to-back, mirroring how a deployed model would actually be trained on a stream of incoming tasks.

Figure~\ref{fig:cl} shows evaluation on all three tasks at different points across the full 600-step training run, normalized per task so that $0$ is the base model's accuracy and $100\%$ is the peak reached on the task by either method. \methodname{} reaches near-peak in every stage while learning faster within each stage, mirroring the data-efficiency gap of Section~\ref{sec:de} Advantage 1. The contrast is sharpest in the second stage, \texttt{CodeIO}: across the full 200-step budget, RL barely lifts off its starting accuracy, peaking at $21.1\%$ mean@4 (a $+2.6$pp gain over its $18.5\%$ stage-start), while \methodname{} climbs to near-peak in about $80$ steps (less than half the budget) and finishes the stage at $39.8\%$, a $+21.3$pp gain (a $\sim$$8\times$ within-stage acquisition rate over RL, and a $+18.7$pp absolute lead at step 400). This demonstrates that \methodname{} is a promising continual-learning algorithm for LLMs: by routing task-level adaptation through both the textual fast-weight channel $\Phi$ in addition to $\theta$, the method remains capable of acquiring later tasks under continued optimization. Appendix~\ref{app:cl-baselines} compares \methodname{} against additional continual-learning baselines, including experience replay, KL regularization, and task-boundary prompt optimization. Appendix~\ref{app:sec:cl} also reports a second task ordering and a longer stream of smaller tasks.


\vspace{-1mm}
\section{Why Does Fast-Slow Training Work?}
\label{sec:ablations}
\vspace{-2mm}
The empirical benefits in Section~\ref{sec:benefits} raise the questions: where do the benefits come from exactly and which component is doing the majority of the work in which setting?
The two studies below isolate these questions.

\paragraph{Observation 1: Fast Weights Acquire Task Signal Faster Than Slow Weights}
\label{sec:abl:star-graph}

\begin{figure}[t]
    \centering
    \includegraphics[width=\linewidth]{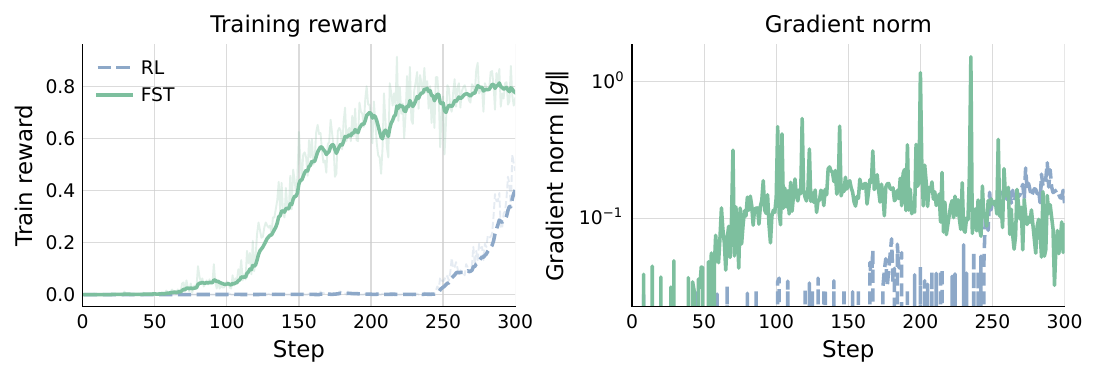}
    \vspace{-5mm}
    \caption{\textbf{Star Graph Search Task.} \textbf{Left:} Training reward improves around step $50$ for \methodname{}, versus $250$ for RL. \textbf{Right:} Gradient norm $\lVert g\rVert$ (log scale) shows earlier sustained gradient signal for \methodname{}'s slow-weight updates; RL's gradients remain mostly small until reward improves.}
    \label{fig:star-graph}
\end{figure}

We compare \methodname{} and RL on a synthetic star-graph task with near-zero base-model reward: finding a path between two labeled nodes (Appendix~\ref{app:star-graph}).

Parameter-only RL remains near zero for roughly $250$ steps, whereas \methodname{} reaches measurable reward around step~$50$ (Figure~\ref{fig:star-graph}), driven primarily by the first few GEPA cycles. Textual feedback identifies where a submitted path fails, allowing GEPA to extract task structure from a handful of rollouts and encode it in $\Phi$ before substantial slow-weight adaptation. Although these early prompts solve only a few problems, they provide an initial reward signal that enables subsequent RL updates to improve rapidly.

\paragraph{Observation 2: Fast and Slow Weights Both Optimizing for Reward Raise Performance Ceiling}
\label{sec:abl:explicit-distill}

\begin{figure}[t]
    \centering
    \includegraphics[width=\linewidth]{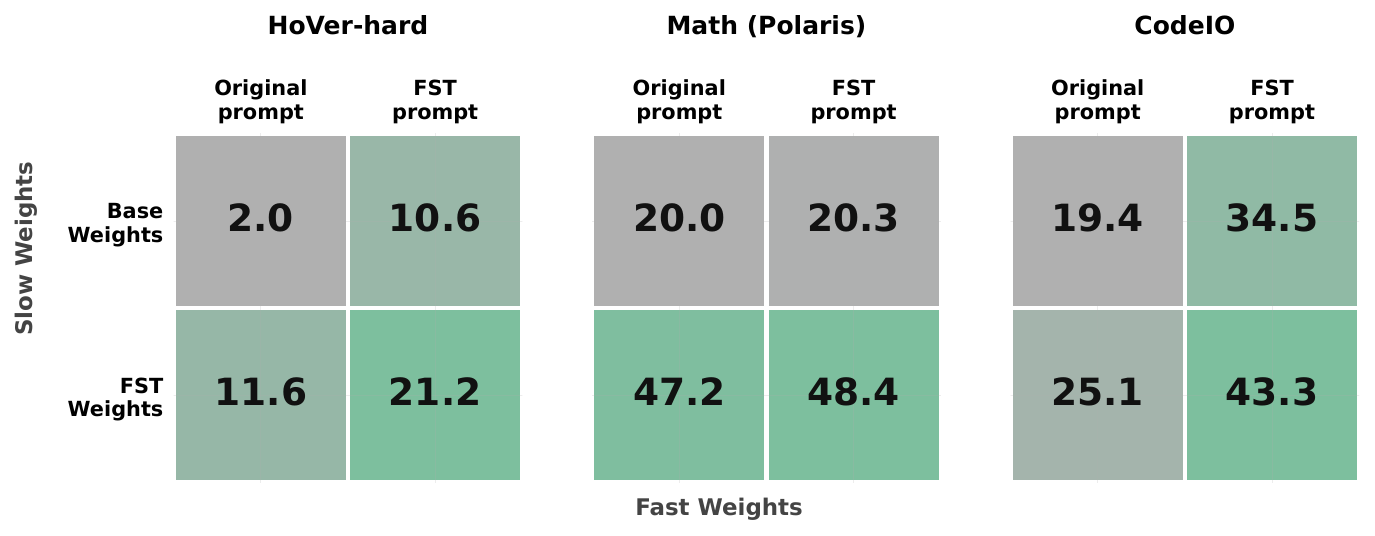}
    \vspace{-1em}
    \caption{\textbf{In-distribution gain decomposed into slow- and fast-weight contributions} (pass@1, \%). For each task, we evaluate four combinations: base or \methodname{}-trained model weights, with the original prompt or the \methodname{}-evolved prompt. On \texttt{HoVer-hard} and \texttt{CodeIO}, both channels contribute and the joint cell (\methodname{} weights + \methodname{} prompt) dominates. On \texttt{Math (Polaris)}, almost all of the gain is carried by the slow weights.}
    \label{fig:indist-quadrants}
\end{figure}

Figure~\ref{fig:indist-quadrants} decomposes the in-distribution gain on each training task into slow-weight and fast-weight contributions, evaluating every combination of \{base, \methodname{}-trained weights\} with \{original prompt, \methodname{}-evolved prompt\}. On \texttt{HoVer-hard}, combining both channels yields $21.2\%$ pass@1, versus $11.6\%$ with trained model weights alone and $10.6\%$ with evolved prompts alone. \texttt{CodeIO} shows the same complementarity. In contrast, almost all the \texttt{Math (Polaris)} gain comes from slow weights ($20.0\% \to 47.2\%$), consistent with the custom SFT base's weaker instruction-following (Appendix~\ref{app:polaris-kl}). Thus, \methodname{} allows a task-dependent division of labor rather than requiring either channel to dominate. Further decomposition and design ablations appear in Figure~\ref{fig:fast-slow-decomp} and Appendix~\ref{app:codeio-ablations}. Naive fast-to-slow distillation also plateaus below \methodname{} (Appendix~\ref{app:explicit-distill}), supporting direct reward optimization of both channels rather than replacing slow-weight RL with distillation.

\section{Discussion}
\label{sec:discussion}

In Sections~\ref{sec:benefits} and~\ref{sec:ablations}, we describe several benefits of training reasoning models with fast-slow updates. As models with finite capacity are trained across ever more diverse sets of environments, we argue that not all task-specific information need be distilled in the weights of the model. We observe some encouraging properties of the new paradigm. First and foremost, \methodname{} maintains proximity to the base model, enabling a set of features suitable to continual learning: plasticity and lack of forgetting. Secondly, the framework allows for data efficient learning, in part due to the ability to learn from text feedback in the context update, overcoming the widely accepted 1-bit-per-episode information limit of binary RLVR. Finally, we observe healthy diversity during training due to a wide prompt pool. The distinction between context and weight optimization represents a broader split between declarative and procedural knowledge, an important distinction for any general-purpose reasoner.

\methodname{} distributes task adaptation between textual context and model parameters. The results in Sections~\ref{sec:benefits} and~\ref{sec:ablations} suggest that learning from text feedback can reduce model weight displacement needed for adaptation, while preserving plasticity for subsequent tasks.

\subsection{Limitations and future work}
\label{sec:limitations}


Our experiments use CISPO and GEPA. The broader framework permits other slow-weight optimizers, including SFT and preference optimization, and fast-weight optimization of context, tools, or a text/code-based agent harness. Evaluating these broader combinations are a fruitful area for future work. 
Future work should also examine more efficient trajectory reuse and extend the preliminary distillation study in Appendix~\ref{app:explicit-distill}.


\vspace{-1mm}
\section{Related Work}
\label{sec:related}
\vspace{-1mm}


\paragraph{Slow learning: RL for LLM reasoning.}
Verifiable-reward LLM post-training writes every improvement into the model parameters via policy-gradient methods such as PPO, DPO, GRPO, and CISPO~\citep{ppo,dpo,deepseekmath,minimaxm1}, used in most reasoning-RL pipelines~\citep{tulu3,scalerl,qwen3}. Prolonged parametric adaptation shrinks output entropy, raises KL to the base policy, and erodes the model's ability to absorb new tasks, called the \emph{plasticity loss} phenomenon~\citep{Dohare2024,understandingpreventingcapacityloss,Frati_2024,mitigatingplasticitylosscontinual,shenfeld2025rlrazor}. We share this diagnosis but add a fast textual channel that absorbs much of the task-specific adaptation the slow weights would otherwise carry.

\paragraph{Fast learning: prompt and context optimization.}
A parallel literature shows that substantial behavioral gains can come from editing the textual context alone, via discrete-prompt search~\citep{shin2020autoprompt,wen2023pez,prasad2023grips}, LLM-driven prompt proposers~\citep{ape,opro,pryzant2023apo,deng2022rlprompt}, evolutionary methods~\citep{fernando2023promptbreeder,guo2024evoprompt,agarwal2024promptwizard}, compound LM programs~\citep{dspy,miprov2,sordoni2023dln,yuksekgonul2024textgrad,cheng2024trace,wu2025optimas}, evolving agent context~\citep{suzgun2025dynamiccheatsheet,wang2024agentworkflowmemory,xu2025amem,zhang2025ace}, and reflective self-feedback~\citep{reflexion,selfrefine,gepa}. We use GEPA, which maintains a per-instance Pareto frontier of candidate prompts. All these methods are typically applied post-hoc to a frozen checkpoint, leaving the slow and fast channels disjoint in time.

\paragraph{Fast and slow weights: complementary learning systems.}
The fast/slow decomposition predates deep learning, with roots in the neuroscience of complementary learning systems~\citep{mcclelland1995cls,kumaran2016cls} and a long line of fast-weight architectures and dual-timescale learners in neural networks~\citep{hinton1987using,schmidhuber1992learning,ba2016usingfastweightsattend,schlag2021lineartransformerssecretlyfast,anthony2017thinking,pham2022dualnet}. We adopt this framework for LLM post-training, instantiating the fast channel as an evolving population of textual prompts and the slow channel as the model parameters.

\paragraph{Modern fast--slow methods for LLM RL.}
A small but growing body of work combines textual feedback with reward-driven weight updates. BetterTogether~\citep{soylu2024bettertogether} alternates SFT with prompt optimization over a DSPy pipeline, albeit instantiated with prompt optimizers that don't use textual feedback; we extend this to a new training paradigm instantiated in verifiable-reward RL in which a Pareto-frontier prompt population created with textual feedback co-evolves with the policy. LANPO~\citep{lanpo2025} interleaves language and numerical feedback via per-instance reflections; we instead maintain a cross-problem Pareto-frontier population. Recent work E-SPL \citep{zhang2026evolutionarypromptlearningreinforcement} explores prompt optimization and RL at a smaller scale with focus on performance rather than adaptation. mmGRPO~\citep{mmgrpo2025} runs prompt optimization once and then RL on a DSPy program; we interleave the two across cycles. Appendix~\ref{app:static-gepa-rl} compares \methodname{} with this sequential pipeline and with LANPO on CodeIO and HoVer-hard. POPE~\citep{qu2026popelearningreasonhard} prefixes hard prompts with partial reference solutions. By contrast, \methodname{} learns task-level prompts that condition any rollout, and combining the two is a natural direction for future work.

\section{Conclusion}
\label{sec:conclusion}
We present a fast-slow framework for LLM post-training that jointly optimizes the slow model parameters $\theta$ via RL and a fast textual-context population $\Phi$ via reflective prompt evolution, interleaving the two channels. Across CodeIO, Math, and HoVer-hard, this co-optimization reaches matched performance with $1.4$–$3\times$ fewer optimizer steps, attains a higher asymptote, and incurs lower KL displacement than RL alone, which in turn translates into preserved plasticity and stronger continual-learning behavior on new tasks.
More broadly, our results suggest that effective post-training should not ask model parameters to absorb all forms of adaptation. Fast textual weights can capture task-specific and rapidly evolving improvements, while slow weights can focus on consolidating persistent behavior. This division of labor offers a path toward post-training methods that are more data-efficient, less destructive, and more amenable to continual learning.

\section{Acknowledgements}
\label{sec:acknowledgements}
The authors acknowledge the gracious support from the Furiosa AI, Apple, NVIDIA, Macronix, Mozilla team, Open Philanthropy / Coefficient Giving, and Amazon Research. Furthermore, we appreciate the support from Google Cloud, the Google TRC team Prof. David Patterson, along with support from Google Gemini team, and Divy Thakkar. Lakshya A Agrawal is also supported by a Laude Slingshot grant and Laude residency provided by the Laude Institute and an Amazon AI PhD Fellowship. We would like to thank Harman Singh, Nishanth Anand and Reza Bayat for useful discussions related to continual learning. Finally, the authors would also like to thank Josh Sirota and the Eragon team for infrastructure and compute support. Our conclusions
do not necessarily reflect the position or the policy of our sponsors, and no official endorsement should be inferred.

\newpage

\bibliographystyle{plainnat} 
\bibliography{reference}

\clearpage
\appendix
\addtocontents{toc}{\protect\setcounter{tocdepth}{2}}
{\renewcommand{\contentsname}{Appendix contents}\tableofcontents}
\clearpage
\FloatBarrier
\section{Ablations}
\label{app:sec:ablations}

\subsection{CodeIO design ablations}
\label{app:codeio-ablations}

In this section, we ablate the design choices for \methodname{}. We run sweeps on CodeIO (Qwen3-8B
thinking, light-recipe defaults from Appendix~\ref{app:hyperparams})
and reported at a matched RL step ($500$) on the held-out CodeIO
mean@$4$. The unmodified RL-only baseline at the same step
($\mathbf{39.65\%}$) is included in every panel for reference.

\begin{figure}[!htb]
    \centering
    \includegraphics[width=\linewidth]{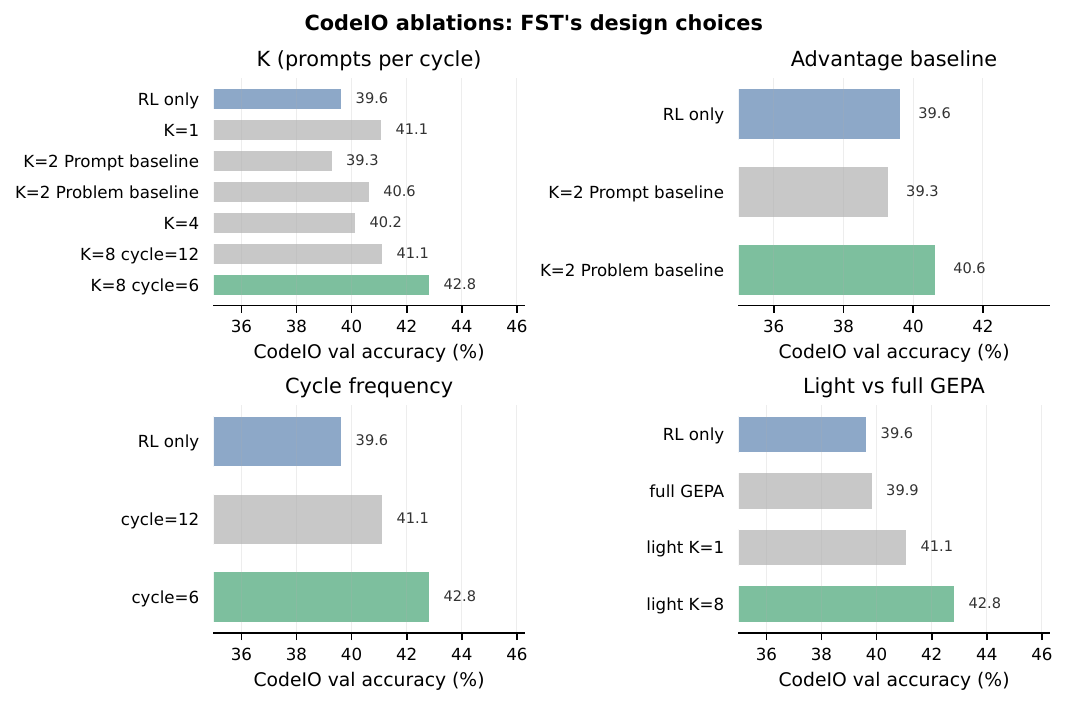}
    \caption{CodeIO design ablations (val mean@$4$ at training step $500$).
    Sage bars mark the headline configuration in each panel; gray bars are
    the alternatives swept; the dashed line indicates RL-only at the same
    matched step. (\textbf{a}) Population size $K$. (\textbf{b}) Advantage
    baseline at $K{=}2$: per-prompt (Prompt baseline) vs.\ per-problem (Problem baseline).
    (\textbf{c}) Cycle length $T$ at $K{=}8$, Problem baseline. (\textbf{d}) Light
    vs.\ full GEPA recipe.}
    \label{fig:codeio-ablations}
\end{figure}

\paragraph{Population size $K$ (Fig.~\ref{fig:codeio-ablations}a).}
We sweep $K \in \{1, 2, 4, 8\}$ holding the rest of the recipe at light /
Problem baseline / cycle $T{=}6$. Every $K \geq 1$ improves on RL-only ($39.65\%$),
indicating that even a single optimized prompt carries useful task signal
into RL. Performance is non-monotonic in $K$: $K{=}1$ already buys
$+1.5$\,pp ($41.10\%$), $K{=}2$ Problem baseline and $K{=}4$ are roughly tied at
$\sim$$40.4\%$, and the gain saturates at $K{=}8$ with $\mathbf{42.84\%}$
(cycle $6$), the headline configuration. Larger $K$ widens the
within-group prompt distribution and gives GRPO advantages a richer
signal to compare against, but only when paired with a tight cycle (see
panel~c). $K{=}8$ with cycle$=12$ recovers most but not all of the gain
($41.13\%$).

\paragraph{Advantage baseline at $K{=}2$ (Fig.~\ref{fig:codeio-ablations}b).}
With a fixed population of $K{=}2$ prompts, the choice of GRPO advantage
baseline matters. The \emph{Prompt baseline} (per-prompt: rollouts under each
$\phi^{(k)}$ are normalized against their own group mean and std) yields
$39.30\%$, slightly \emph{below} RL-only. The \emph{Problem baseline}
(per-problem: all $G$ rollouts under both prompts share a single group
statistic) yields $\mathbf{40.65\%}$, $+1.4$\,pp over the Prompt baseline
and $+1.0$\,pp over RL-only. Our intuition for the problem baseline working well is as follows. The prompt baseline makes
prompts compete only with themselves and discards the cross-prompt
comparison entirely, while the problem baseline exposes the policy
gradient to which prompt a stronger response came from on the same
problem. The problem baseline is the default for all main-text experiments.

\paragraph{Cycle length $T$ at $K{=}8$ (Fig.~\ref{fig:codeio-ablations}c).}
Holding $K{=}8$ Problem-baseline fixed, we compare $T{=}6$ vs.\ $T{=}12$ RL steps
between successive GEPA optimizations. Cycle $T{=}6$ reaches
$\mathbf{42.84\%}$; doubling the cycle to $T{=}12$ drops mean@$4$ by
$1.7$\,pp to $41.13\%$, giving up more than half of the advantage over
RL-only. This is the expected effect of staleness: as $\theta$ moves between
GEPA cycles, the prompts in $\Phi$ become increasingly mistuned to the
current policy, and the per-question rollout-group signal degrades.
$T{=}6$ is short enough that the population $\Phi_c$ remains close to
optimal across the cycle, though it requires twice as many GEPA optimizations as cycle 12.

\paragraph{Light vs.\ full GEPA recipe (Fig.~\ref{fig:codeio-ablations}d).}
The ``light'' recipe uses $K{=}4$ candidates, a smaller per-cycle GEPA
budget (\texttt{num\_eval\_examples}$=192$,
\texttt{max\_metric\_calls}$=960$), and a proposer prompt that asks
\texttt{gpt-5.2} for incremental edits to the current best prompt rather
than rewrites from scratch. The ``full'' recipe is the original
configuration from~\citet{gepa}: $K{=}1$, doubled metric budget
(\texttt{max\_metric\_calls}$=1922$), and an open-ended proposer that
allows full rewrites. Within our setup the full recipe gives essentially
no lift over RL-only ($39.85\%$ vs.\ $39.65\%$); the light recipe at the
same $K{=}1$ yields $+1.5$\,pp ($41.10\%$); and scaling light to $K{=}8$
gives a further $+1.7$\,pp ($\mathbf{42.84\%}$). We believe two factors contribute.
First, full's $K{=}1$ pinches the population channel that turns out to
matter most in panel~a. Second, the open-ended rewrite proposer is more
prone to drift on a moving policy: incremental edits keep the
population's induction biases coherent across cycles, while wholesale
rewrites tend to discard structure each round. The combined effect is a
recipe that runs on roughly half the GEPA budget and outperforms the
original by $\sim$$3$\,pp on this task.

\subsection{A wider cycle-length sweep}
\label{app:cycle-sweep}

\begin{figure}[!htb]
    \centering
    \includegraphics[width=\linewidth]{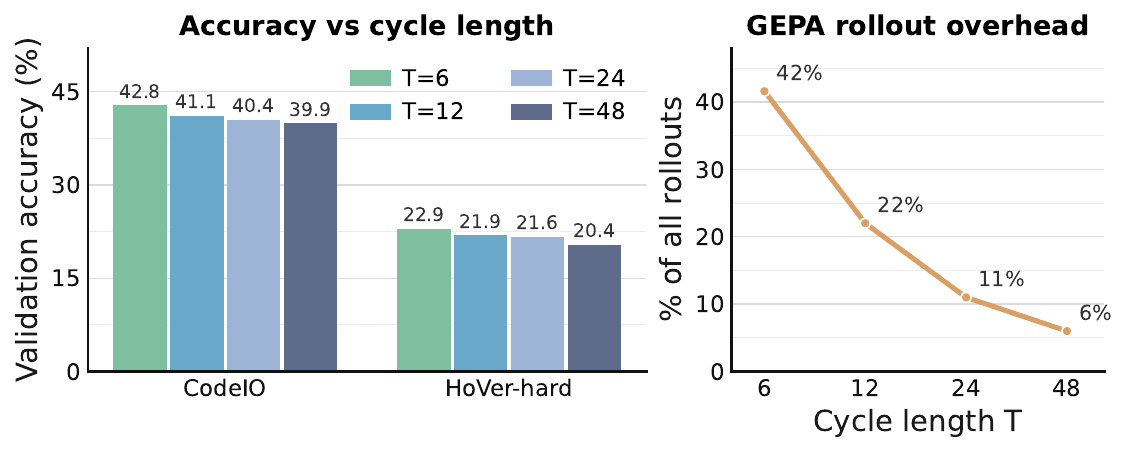}
    \caption{\textbf{Accuracy is weakly sensitive to cycle length $T$; rollout
    cost is strongly sensitive.} \emph{Left:} held-out accuracy for
    $T \in \{6, 12, 24, 48\}$ on CodeIO (mean@$4$, $700$ training steps) and
    HoVer-hard (mean@$4$, $300$ training steps). \emph{Right:} GEPA rollouts as
    a share of all rollouts generated during training, for the same runs.}
    \label{fig:cycle-length-wide}
\end{figure}

Panel~c compares $T{=}6$ against $T{=}12$ on a single task. To check that
longer cycles do not eventually pay off, we extend the sweep to
$T \in \{6, 12, 24, 48\}$ and repeat it on HoVer-hard. Accuracy declines
monotonically but slowly with $T$: from $42.8\%$ at $T{=}6$ to $39.9\%$ at
$T{=}48$ on CodeIO, and from $22.9\%$ to $20.4\%$ on HoVer-hard. Cost moves far more over the
same range: the share of rollouts spent by GEPA falls from $41.6\%$ at
$T{=}6$ to $6.0\%$ at $T{=}48$. Short cycles keep the prompt population
tightly coupled to the current policy and buy a small, consistent accuracy
edge; long cycles give most of that edge back in exchange for a several-fold
reduction in prompt-optimization overhead. In practice we therefore pick $T$
by rollout budget: $T{=}6$ when rollouts are cheap, larger $T$ when they are
not, with the expectation of losing roughly a point of accuracy per doubling.

\subsection{Replication on HoVer-hard}
\label{app:hover-ablations}

\begin{figure}[!htb]
    \centering
    \includegraphics[width=\linewidth]{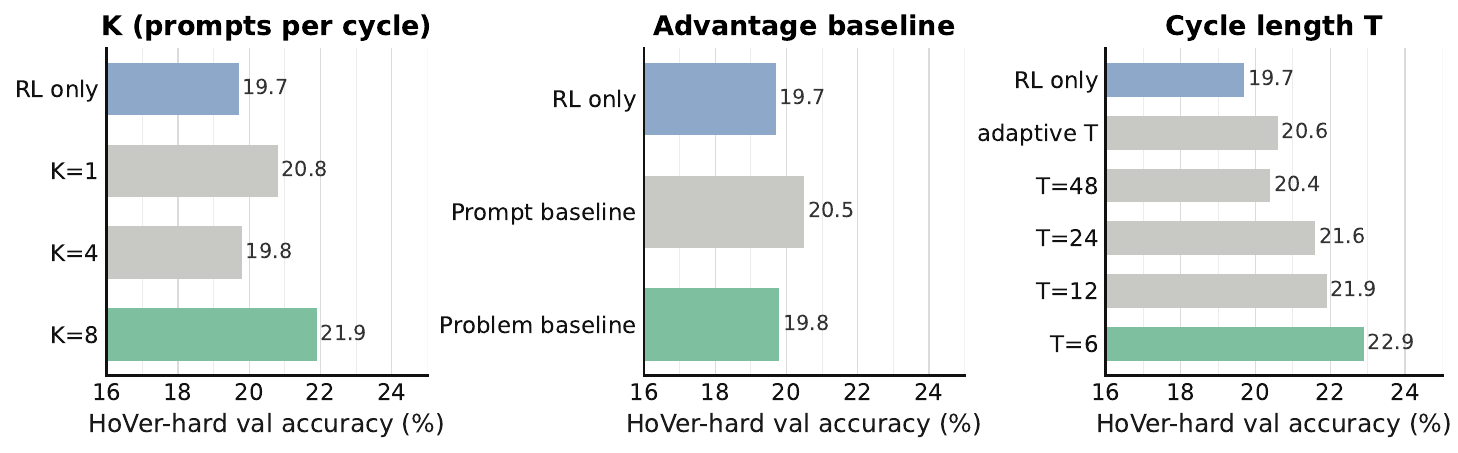}
    \caption{\textbf{Design ablations replicated on HoVer-hard} (held-out mean@$4$ at the best checkpoint of a $300$-step run). Same layout as Figure~\ref{fig:codeio-ablations}: sage marks the headline configuration, grey the alternatives, blue the RL-only run. (\textbf{Left}) Population size $K$. (\textbf{Middle}) Advantage baseline. (\textbf{Right}) Cycle length $T$, including an adaptive rule that starts a GEPA cycle only when the training reward stalls.}
    \label{fig:hover-ablations}
\end{figure}

The CodeIO sweeps above are single-task, so we repeated the three component ablations on HoVer-hard. The population-size trend replicates: $K{=}8$ reaches $21.9\%$ against $20.8\%$ for a single prompt and $19.8\%$ for $K{=}4$. The cycle-length trend replicates too, and it is the largest effect on this task: accuracy falls from $22.9\%$ at $T{=}6$ to $20.4\%$ at $T{=}48$. The advantage baseline is the one component whose effect does not carry over: on HoVer-hard the per-prompt baseline ($20.5\%$) and the per-problem baseline ($19.8\%$) are within a point of each other, whereas on CodeIO the per-prompt baseline fell below RL-only. We keep the per-problem baseline throughout the paper because it never hurts and is decisive on CodeIO. Finally, to check that $T$ need not be hand-tuned, we ran a variant that starts a GEPA cycle only when an exponential moving average of the training reward has stopped improving. It reaches $20.6\%$, between the $T{=}24$ and $T{=}48$ runs, at a fraction of their prompt-optimization cost.

\FloatBarrier
\section{Baselines}
\label{app:sec:baselines}

\subsection{Prompt optimization once, then RL}
\label{app:static-gepa-rl}

\begin{figure}[!htb]
    \centering
    \includegraphics[width=\linewidth]{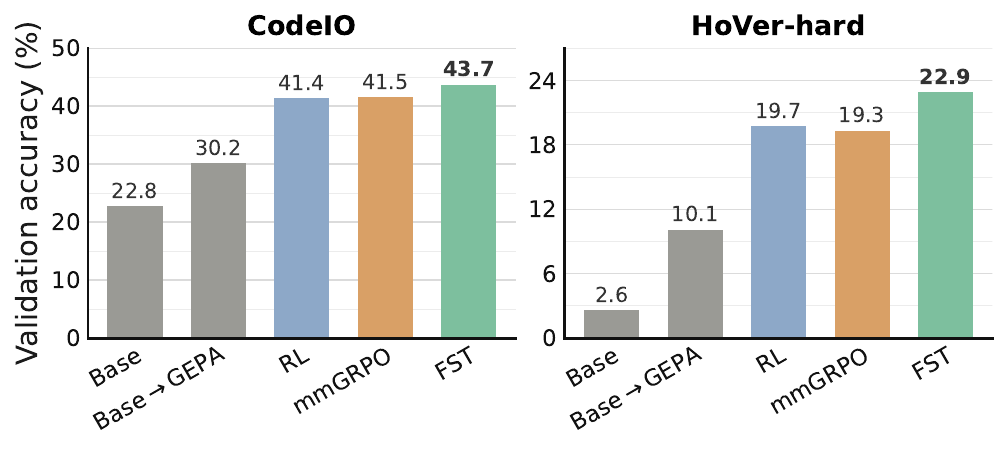}
    \caption{\textbf{A strong static prompt does not reproduce \methodname{}.}
    Held-out accuracy (mean@$4$) for the base model, the base model with a
    GEPA-optimized prompt, RL, the static pipeline (GEPA once on the base
    model, prompt frozen, then RL; the recipe of
    mmGRPO~\citep{mmgrpo2025}), and \methodname{}. CodeIO values are compared
    at a matched training step ($1100$); HoVer-hard values are the best
    checkpoint per arm.}
    \label{fig:static-gepa-rl}
\end{figure}

\methodname{} interleaves prompt optimization with RL. A natural alternative,
and the recipe of mmGRPO~\citep{mmgrpo2025}, is to optimize the prompt once on
the base model, freeze it, and run RL on top. Comparing the two separates the
benefit of \emph{co-evolving} the prompt with the policy from the benefit of
merely having a strong prompt available during training. We run this pipeline
on CodeIO and HoVer-hard: GEPA on the base model, then the unmodified RL
baseline with the best resulting prompt held fixed throughout training.

Figure~\ref{fig:static-gepa-rl} shows the result. On both tasks the static
pipeline lands on the RL baseline rather than between RL and \methodname{}:
$41.5\%$ versus $41.4\%$ on CodeIO, and $19.3\%$ versus $19.7\%$ on
HoVer-hard, against $43.7\%$ and $22.9\%$ for \methodname{}. The GEPA
prompt with used with the base model lifts CodeIO
from $22.8\%$ to $30.2\%$ and HoVer-hard from $2.6\%$ to $10.1\%$. However, through training RL eventually learns the same task knowledge and both arms arrive at the same place. Optimizing the prompt once and training
cannot reproduce is the co-adaptation of prompt and weights during training,
which is where \methodname{}'s remaining $2$--$3$\,pp comes from.

\begin{figure}[!htb]
    \centering
    \includegraphics[width=\linewidth]{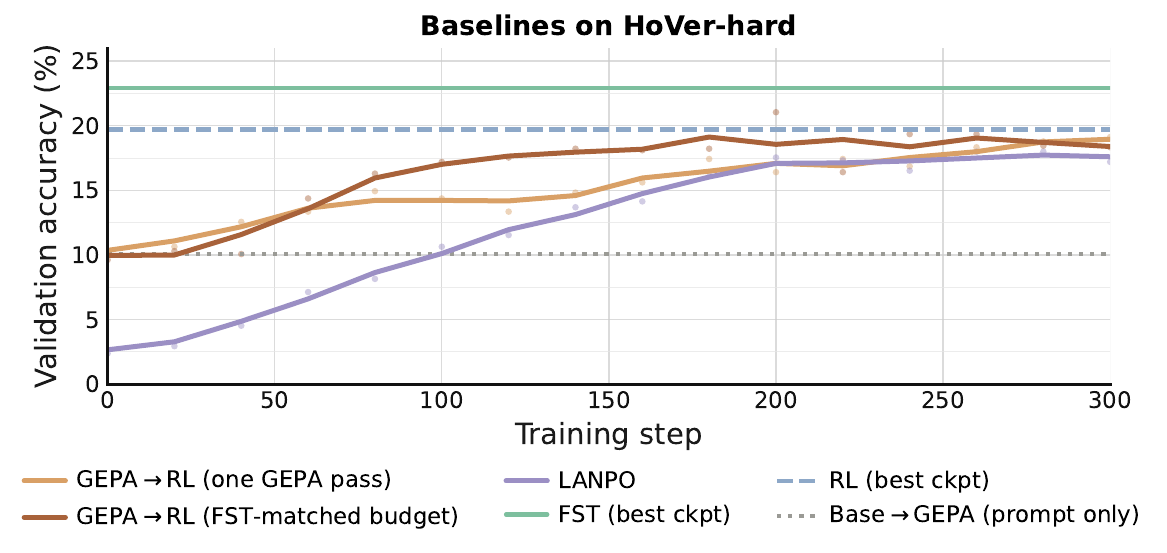}
    \caption{\textbf{Baselines on HoVer-hard} (held-out mean@$4$, evaluated every $20$ steps). Two GEPA$\rightarrow$RL arms differ only in the budget of the GEPA stage: one GEPA pass, or the metric-call budget that \methodname{} spends over a whole $300$-step run. Horizontal lines mark the best checkpoints of RL and \methodname{} from Figure~\ref{fig:static-gepa-rl} and the prompt-only level. Both static-prompt arms and LANPO~\citep{lanpo2025} finish at or below RL and well below \methodname{}.}
    \label{fig:hover-sequential-curves}
\end{figure}

\subsection{More prompt-optimization budget for the static pipeline}
\label{app:matched-budget}

One objection to the comparison above is that \methodname{} runs GEPA every $T$ steps while the static pipeline runs it once, so the static prompt might simply be under-optimized. We therefore reran the pipeline on HoVer-hard with the GEPA stage given the same metric-call budget that \methodname{} consumes over a full $300$-step ru.
Figure~\ref{fig:hover-sequential-curves} shows RL on top of the two prompts, alongside LANPO (next paragraph). At step $300$ the standard-budget arm is at $18.7\%$ and the matched-budget arm at $18.5\%$, against $22.9\%$ for \methodname{}. The two arms differ only in how much prompt optimization is done before RL, and they end in the same place, so the gap to \methodname{} comes from \emph{when} the prompt optimization is spent, during RL against the current policy, rather than from how much of it there is.

\subsection{Prompt optimization after RL}
\label{app:rl-then-gepa}

The other sequential order is RL first and GEPA afterwards on the trained checkpoint. Evaluating all arms under one protocol (pass@$1$ over $8$ samples on the $221$ HoVer-hard test problems, the protocol of Figure~\ref{fig:indist-quadrants}), RL alone reaches $16.4\%$, and running GEPA on the RL checkpoint at evaluation raises it to $19.3\%$. \methodname{}, evaluated with its own evolved prompt, reaches $21.2\%$. Handing the model weights from \methodname{} a prompt that was instead optimized for the base model \emph{lowers} their accuracy to $17.9\%$: the weights and the prompt were adapted to each other during training, and neither order of sequential optimization reproduces that pairing.

\subsection{LANPO}
\label{app:lanpo}

LANPO~\citep{lanpo2025} also feeds language back into RL, but through a different channel: before each rollout, the policy is shown a reflection over its own earlier attempts at the same problem, drawn from a per-problem experience pool, with no prompt optimizer and no external reflection model. We implemented its intra-sample variant (context assigned per problem so that GRPO groups stay comparable, $p{=}0.75$ probability of including the reflection, and the format bonus of the original) and trained it with the HoVer-hard RL recipe for $300$ steps. On HoVer-hard, LANPO learns steadily from the base model's $2.4\%$ and reaches a best checkpoint of $18.0\%$ ($17.2\%$ at step $300$), below RL-only ($19.7\%$) and $5$\,pp below \methodname{} (Figure~\ref{fig:hover-sequential-curves}). On CodeIO it reaches $37.9\%$, against $41.4\%$ for RL-only and $43.7\%$ for \methodname{}. Reflecting on the policy's own rollouts therefore does not substitute for optimizing the prompt against reward.

\FloatBarrier
\section{Efficiency}
\label{app:sec:efficiency}

\subsection{Rollout reuse}
\label{app:rollout-reuse}

\begin{figure}[!htb]
    \centering
    \includegraphics[width=\linewidth]{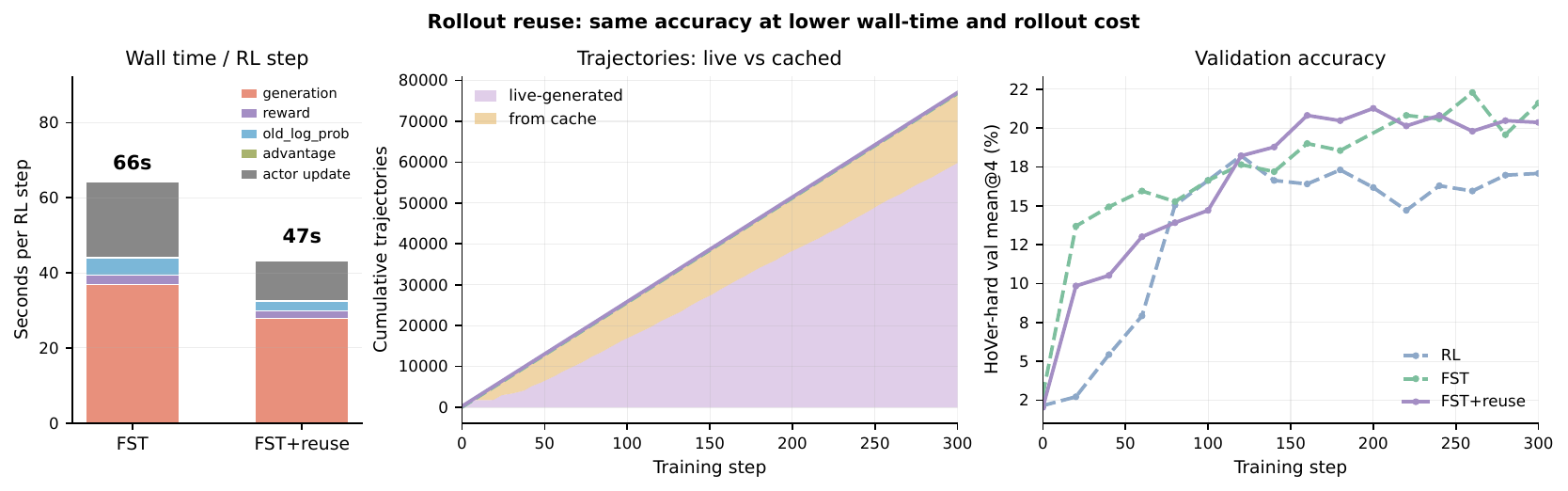}
    \caption{Rollout reuse on HoVer-hard, training step $\le$300.
    \textbf{Left:} mean wall-time per RL step. \methodname{}+reuse drops from $\sim$66\,s
    to $\sim$47\,s (about $30\%$ faster); the saving comes almost entirely from
    the generation phase.
    \textbf{Middle:} cumulative RL trajectories. The shaded region splits each
    step's $256$ trajectories into live-generated (lavender) and from-cache (amber).
    By step 300 about $\sim$17{,}000 of the $\sim$77{,}000 RL trajectories were
    served from cache.
    \textbf{Right:} HoVer-hard val accuracy mean@4. \methodname{}+reuse tracks
    \methodname{}; the no-prompt RL baseline lags both.}
    \label{fig:rollout-reuse}
\end{figure}

\paragraph{Reusing GEPA rollouts for RL.} \methodname{} interleaves GEPA prompt
optimization with RL updates: every $T$ RL steps GEPA runs a cycle that
scores each candidate prompt $\phi^{(k)}$ on a small evaluation pool drawn
from the same training distribution the next RL phase will see. Each evaluation entry is a tuple
$(p, \phi^{(k)}, y, r)$ -- a problem $p$, the prompt $\phi^{(k)}$ used,
the sampled response $y$, and its scalar reward $r$. Without reuse, those
tuples are thrown away once GEPA picks a new population $\Phi_{c+1}$;
the next RL step then re-rolls $G$ fresh trajectories per problem from
scratch under the new prompts. But because the population $\Phi_{c+1}$
is a Pareto-frontier subset of the prompts GEPA \emph{already evaluated},
many of those discarded tuples are perfectly valid samples from the
current policy under one of the current prompts. The natural optimization
is to splice them into the RL group instead of regenerating them.

\paragraph{Cache mechanics.} We maintain a per-(problem, prompt) cache of
GEPA evaluation tuples produced during the most recent cycle. When the RL
phase forms its rollout group of size $\text{batch} \times G$ across the
prompts in $\Phi_{c+1}$, it first claims any cached $(p, \phi, y, r)$
matching a (problem, prompt) slot and only generates the remaining slots
live with vLLM. Cached and live trajectories are then concatenated into a
single GRPO group; the policy gradient does not distinguish the two
sources because both were sampled under prompts in the active population.
The cache is cleared when GEPA produces the next $\Phi_{c+2}$, so reused
trajectories are at most $T$ RL steps old (
$T{=}6$ in our headline configuration).

\paragraph{Empirical effect.} Figure~\ref{fig:rollout-reuse} measures the
impact on HoVer-hard. We compare two otherwise identical \methodname{} runs --
one with reuse enabled and one without -- through the first $300$ training
steps, with a no-prompt RL baseline as a third reference. Wall-time per
RL step (left panel) drops from $\sim$66\,s to $\sim$47\,s, a $\sim$$29\%$
speedup that comes almost entirely out of the generation phase: GEPA's
own per-cycle cost is unchanged, but a substantial fraction of the
following RL phase's rollouts no longer need to be sampled. The middle
panel decomposes the cumulative trajectory budget into the live and
from-cache components: by step $300$, roughly $17$k of the $77$k total
trajectories ($\sim$$22\%$) were served from cache, with the cache hit
rate concentrated in the first few RL steps after each GEPA cycle and
tapering as the live group fills in problems not in GEPA's evaluation
pool. The right panel shows that this comes at no accuracy cost: the
reuse and non-reuse \methodname{} curves are within sampling noise of each
other on HoVer-hard val mean@$4$, and both lead the no-prompt RL
baseline throughout.

\subsection{Total rollout and reflection cost}
\label{app:cost}

The step counts in Section~\ref{sec:de} compare data consumed: one RL step is one batch of $32$ problems with $G{=}8$ rollouts each for both \methodname{} and RL, and GEPA draws its reflection examples from problems the RL loop has already consumed, so a step is the same unit for both methods. GEPA does, however, generate additional rollouts of its own, mostly to score candidate prompts on its validation pool, and we account for those here. Table~\ref{tab:cost} gives the per-step averages over a full run at $T{=}6$. GEPA adds $92$--$123$ rollouts per RL step, an overhead of $18$--$48\%$, and about one to three reflection-LM calls per step (under \$$10$ of API cost per training run). Because \methodname{} reaches RL's peak in $1.4$--$3\times$ fewer steps, its total rollout count at matched accuracy is still lower than RL-only's on every task: $51\%$ fewer on CodeIO and HoVer-hard and $16\%$ fewer on Math. Wall-clock per step is higher and depends on $T$: on CodeIO, RL-only takes $163$\,s per step, \methodname{} takes $329$\,s at $T{=}6$ ($2.0\times$) and $249$\,s at $T{=}12$ ($1.5\times$), before the rollout reuse described above, which on HoVer-hard brings \methodname{}'s per-step wall-clock below RL-only's.

\begin{table}[!htb]
\centering
\small
\caption{Rollout and reflection cost of \methodname{} at $T{=}6$. Rollouts and reflection-LM calls are averaged per RL step over a full run (a GEPA cycle fires every $6$ steps). The last column compares the total number of rollouts each method has generated when it first reaches RL's peak accuracy.}
\label{tab:cost}
\begin{tabular}{@{}lccccc@{}}
\toprule
 & \multicolumn{2}{c}{Rollouts / step} & Reflection & Step & Rollouts at \\
\cmidrule(lr){2-3}
Task & RL & GEPA (overhead) & calls / step & reduction & matched acc. \\
\midrule
CodeIO         & 256 & 116 (45\%) & 0.96 & $3.0\times$ & $-51\%$ \\
HoVer-hard     & 256 & 123 (48\%) & 2.72 & $3.0\times$ & $-51\%$ \\
Math (Polaris) & 512 & \phantom{0}92 (18\%) & 1.70 & $1.4\times$ & $-16\%$ \\
\bottomrule
\end{tabular}
\end{table}

\FloatBarrier
\section{Additional results}
\label{app:sec:additional}

\subsection{KL-vs-reward}
\label{app:polaris-kl}

\begin{figure}[!htb]
    \centering
    \includegraphics[width=\linewidth]{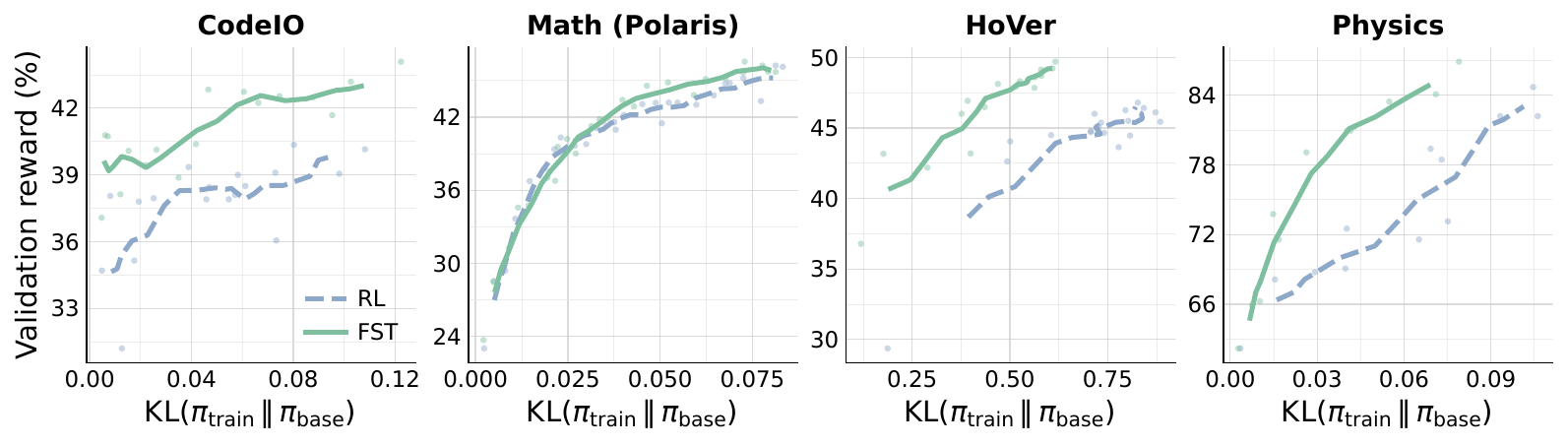}
    \caption{Validation reward versus $\mathrm{KL}(\pi_{\mathrm{train}}\,\|\,\pi_{\mathrm{base}})$ on all four training tasks: \texttt{CodeIO}, \texttt{Math} (Polaris), \texttt{HoVer}, and \texttt{Physics}. Same axes, smoothing, and conventions as Figure~\ref{fig:kl-vs-acc} in the main text. The three-task variant in the main text drops the Polaris panel for the reasons discussed below.}
    \label{fig:kl-vs-acc-full}
\end{figure}

The Polaris (math) trajectories in Figure~\ref{fig:kl-vs-acc-full} look qualitatively different from the other three tasks: RL and \methodname{} sit on top of each other in $(\mathrm{KL}, \text{reward})$-space rather than \methodname{} pulling the frontier to the left. We attribute this to the base model used for the Polaris runs. Unlike the other three tasks, which start from \texttt{Qwen3-8B} (an instruction-tuned model with strong format following), Polaris was trained on top of a model built by SFT'ing \texttt{Qwen3-8B-Base} on Nemotron, since the public \texttt{Qwen3-8B} is already saturated on Polaris. That custom SFT base has noticeably weaker instruction-following than the public Instruct checkpoint: the GEPA-evolved prompts that drive \methodname{}'s KL gain on the other three tasks rely on the policy actually following format and self-checking instructions in the prompt, and on this base much of that signal is lost. The model still learns the math task from RL reward, so the reward axis behaves normally. Tt simply does not benefit from the prompt channel as strongly, and the two trajectories collapse onto each other. We expect the Polaris curve to look more like the other three with a stronger instruction-tuned base, but isolating that requires retraining and is left for future work.

\subsection{Scaling the base model}
\label{app:model-size}

\begin{figure}[!htb]
    \centering
    \includegraphics[width=\linewidth]{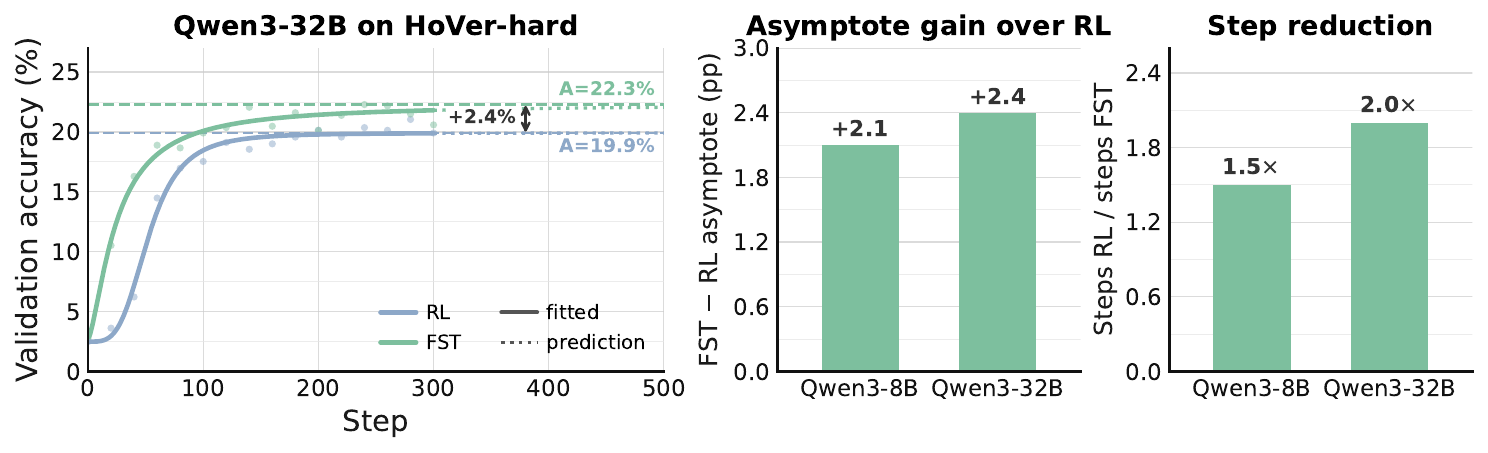}
    \caption{\textbf{\methodname{} scaling: two model sizes on HoVer-hard} ($300$ training steps, $T{=}12$, otherwise the settings of Appendix~\ref{app:hyperparams}). \textbf{Left:} Qwen3-32B validation curves for RL-only and \methodname{}, in the conventions of Figure~\ref{fig:performance-asymptote}: points are the per-evaluation values, the solid line is the fitted sigmoid of Section~\ref{sec:asymptote}, dotted is its extrapolation, and dashed lines mark the fitted asymptotes $A$. \textbf{Middle:} the asymptote of \methodname{} minus that of RL-only at each size. \textbf{Right:} the number of steps RL-only needs to reach its own peak reward divided by the number \methodname{} needs.}
    \label{fig:model-size}
\end{figure}

Our results in the main paper use Qwen3-8B. To check that the gain survives a stronger base model, we trained Qwen3-32B (thinking mode) on HoVer-hard with RL-only and with \methodname{}. The 32B runs are shorter than the 8B headline runs ($300$ steps) and use $T{=}12$, so we compare them against 8B runs with the same budget and cycle length, and we read the comparison off the fitted asymptote of each arm rather than off the best checkpoint, which also removes the effect of where each run was stopped. Figure~\ref{fig:model-size} shows the 32B curves and the two quantities that carry the paper's argument. At 32B the fitted asymptotes are $19.9\%$ for RL-only and $22.3\%$ for \methodname{}. The \methodname{} curve is still rising at step $300$, so its fit is floored at the observed maximum and its asymptote is a lower bound. \methodname{} first reaches RL-only's peak ($21.0\%$) at step $140$ against RL-only's step $280$. The asymptote gap between \methodname{} and RL-only is $+2.1$\,pp at 8B and $+2.4$\,pp at 32B, and the step reduction grows from $1.5\times$ at 8B to $2.0\times$ at 32B. At matched reward the 32B \methodname{} checkpoint also sits at $78\%$ lower KL to the base model than the RL-only checkpoint. The raw best-checkpoint accuracies at 32B are $21.0\%$ for RL-only and $22.3\%$ for \methodname{}, from a base of $2.6\%$, the same base accuracy as at 8B, so the task does not become easy with scale. Appendix~\ref{app:nonverifiable} covers the other direction, a smaller model (Qwen3-4B) and a second family (OLMo-3-7B), on a rubric-scored task.

\subsection{Explicit fast-to-slow distillation}
\label{app:explicit-distill}

The ceiling finding in Section~\ref{sec:abl:explicit-distill} raises a natural question: can the gains from the fast textual channel be folded into model parameters explicitly, without ever doing RL on the slow weights? We test this by replacing the slow-weight policy-gradient update with an on-policy reverse-KL distillation loss
\begin{equation}
\label{eq:distill}
\mathcal{L}_{\text{distill}}(\theta) \;=\; \mathbb{E}_{x \sim \mathcal{D},\; y \sim \pi_\theta(\cdot \mid x)} \!\left[\, \sum_{t=1}^{|y|} \mathrm{KL}\!\Big(\pi_\theta(\cdot \mid x, y_{<t}) \,\Big\Vert\, \pi_{\bar{\theta}}(\cdot \mid x, \phi, y_{<t}) \Big) \,\right],
\end{equation}
where the teacher $\pi_{\bar{\theta}}(\cdot \mid x, \phi, \cdot)$ is the same model evaluated with a \methodname{}-evolved fast-weight prompt $\phi$ and fixed model weights $\bar{\theta}$, and the student $\pi_\theta(\cdot \mid x, \cdot)$ is conditioned only on the problem $x$. Sampling $y$ on-policy from the student and minimizing the per-token reverse KL toward the teacher follows recent work on self-distillation~\citep{hubotter2026sdpo}. We call this variant \methodname{}-distill: the slow weights move only by chasing the teacher distribution induced by the fast-weight prompt, with no direct exposure to scalar reward.

\begin{figure}[!htb]
    \centering
    \includegraphics[width=\linewidth]{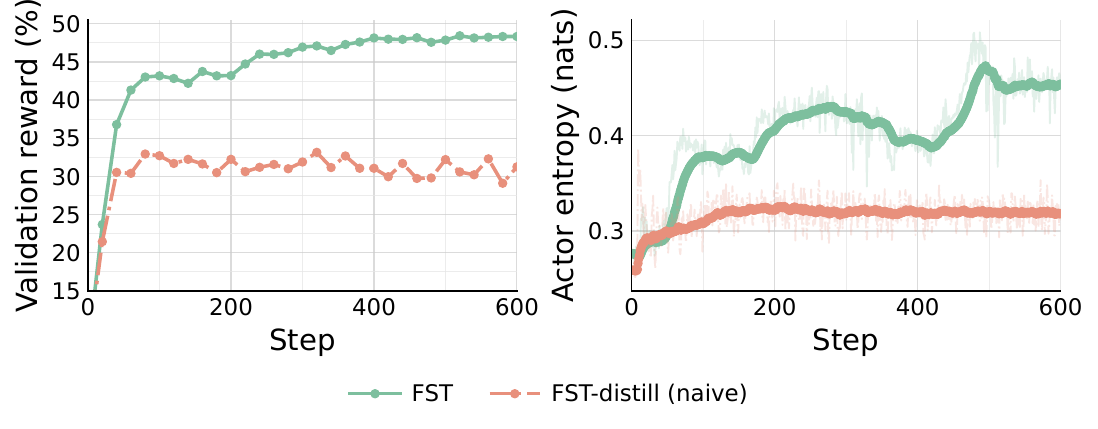}
    \vspace{-1em}
    \caption{\textbf{Explicit fast-to-slow distillation on \texttt{HoVer}.} \methodname{} (green) is compared to \methodname{}-distill (orange), which updates $\theta$ only via the on-policy reverse-KL loss in Eq.~\ref{eq:distill} using a \methodname{}-evolved prompt $\phi$ as the teacher. \textbf{Left:} Validation reward. \methodname{}-distill rises above the prompt-only level by transferring fast-weight signal into model parameters across multiple updates, but plateaus well below \methodname{}, which has both channels optimizing reward jointly. \textbf{Right:} Actor entropy. Both methods preserve healthy entropy throughout training.}
    \label{fig:beyond-fast-weights}
\end{figure}

Figure~\ref{fig:beyond-fast-weights} (left) shows that \methodname{}-distill iteratively transfers signal from the fast weights into the slow weights and rises above the prompt-only ceiling, but does not reach \methodname{}'s reward. This is consistent with Observation~2. The fast channel alone is not enough to saturate the joint ceiling, and direct policy-gradient updates on $\theta$ against reward, run alongside the fast channel, account for the remaining gain. The entropy panel (right) also shows that the diversity benefits of training under a Pareto-frontier prompt population persist even when the slow update is purely distillation-based.

\begin{figure}[!htb]
    \centering
    \includegraphics[width=\linewidth]{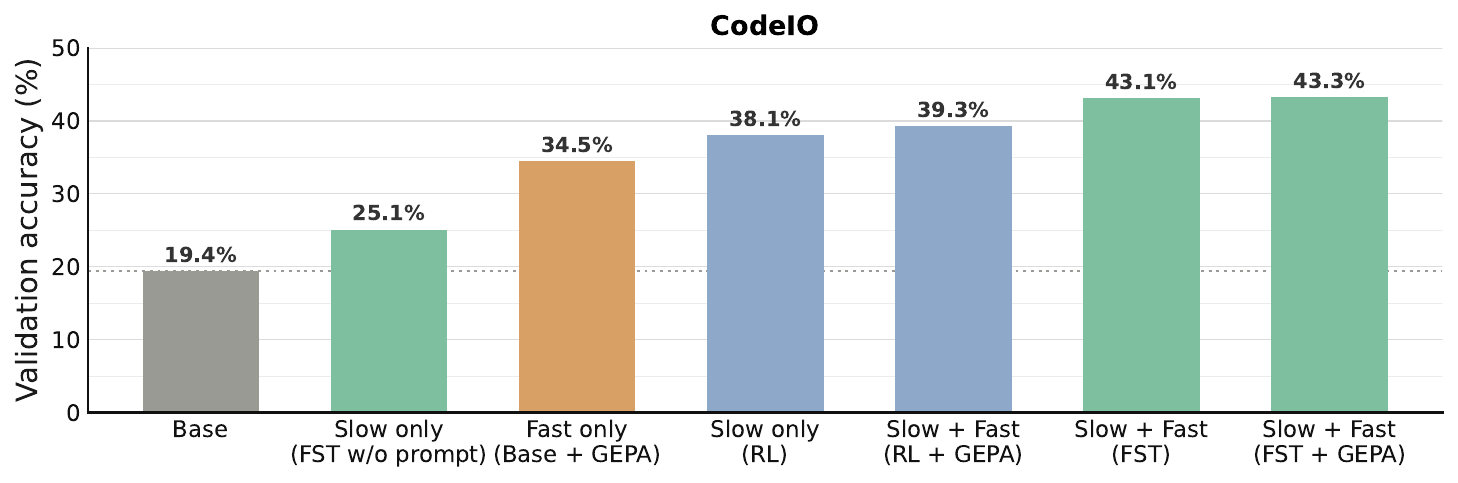}
    \caption{\textbf{Decomposing the Fast vs. Slow gain on CodeIO.} Step-matched (training step $650$) validation accuracy (pass@$1$, computed from $n{=}8$ rollouts) on the held-out CodeIO set. \emph{Slow only} isolates the parametric channel (RL- or \methodname{}-trained weights, evaluated without any GEPA prompt); the RL value is the training-time validation accuracy of the RL-only run at step $640$, its nearest logged evaluation to step $650$. \emph{Fast only} isolates the textual channel (base model weights with a GEPA-evolved prompt). \emph{Slow~+~Fast} combines them. Both channels contribute. \methodname{} compounds the benefits: without its prompt the \methodname{} model weights score $25.1\%$, below RL's $38.1\%$, but with the prompt they were trained with, the full Slow~+~Fast (\methodname{}) configuration reaches $43.1\%$, above RL given a GEPA prompt after training ($39.3\%$), and an extra GEPA pass on top lifts it further to $43.3\%$.}
    \label{fig:fast-slow-decomp}
\end{figure}

\subsection{Evolved GEPA prompts during \methodname{} training}
\label{app:gepa-prompts}

This section shows, for each \methodname{} training task, the
\emph{seed} prompt that GEPA started from and the \emph{evolved}
prompt at the matched-step \methodname{} checkpoint used in §\ref{sec:de}.
The evolved prompt is the population's lead candidate
(\texttt{gepa\_state.json:current\_prompts[0]}) of each Problem-baseline
$K{=}\{4,8\}$ training run at that step, so it is a prompt that was
\emph{co-evolved with the slow-weight RL update}, not a prompt obtained
by running GEPA on the un-trained base policy in isolation.
\methodname{}'s rollout group at that step also draws from the rest of
the $K$-prompt population, not just the single one shown here.

Across tasks, two patterns are consistent. First, GEPA almost never
\emph{rewrites} the seed. The $K{=}\{4,8\}$ Problem-baseline recipe
constrains the proposer (\texttt{gpt-5.2}) to small targeted edits, so
the evolved prompt keeps the seed's basic role and output format and
adds layered guidance on top. Second, the additions are almost entirely
\emph{negative-example specific}: each block addresses a failure mode
the proposer observed during reflection on a small batch of low-reward
rollouts (e.g., ``do not invent placeholder numbers'' for CodeIO,
``do not skip pages with parenthetical disambiguators'' for HoVer-hard,
``re-check off-by-one in process/recurrence problems'' for Polaris).
The result is a long instruction list that reads less like a generic
system prompt and more like a checklist of don'ts assembled from the
specific population of mistakes the policy was making.

\subsubsection{CodeIO}

\promptbox{Seed Prompt -- CodeIO}{gepa_prompts/codeio_seed.txt}{task_codeio}

\promptbox{Evolved Prompt -- CodeIO ($K{=}8$ Problem-baseline, training step 650)}{gepa_prompts/codeio_K8_optB_best_step650.txt}{task_codeio}

\subsubsection{Math (Polaris)}

\promptbox{Seed Prompt -- Math (Polaris)}{gepa_prompts/polaris_seed.txt}{task_polaris}

\promptbox{Evolved Prompt -- Math (Polaris) ($K{=}4$ Problem-baseline, training step 1050)}{gepa_prompts/polaris_K4_optB_best_step1050.txt}{task_polaris}

\subsubsection{Physics}

\promptbox{Seed Prompt -- Physics}{gepa_prompts/physics_seed.txt}{task_physics}

\promptbox{Evolved Prompt -- Physics ($K{=}4$ Problem-baseline, training step 500)}{gepa_prompts/physics_K4_optB_best_step500.txt}{task_physics}

\subsubsection{HoVer-hard}

\promptbox{Seed Prompt -- HoVer-hard}{gepa_prompts/hover_strict_seed.txt}{task_hover}

\promptbox{Evolved Prompt -- HoVer-hard ($K{=}8$ Problem-baseline, training step 550)}{gepa_prompts/hover_strict_K8_optB_best_step550.txt}{task_hover}

\FloatBarrier
\section{Non-verifiable tasks}
\label{app:sec:nonverifiable}

\subsection{RaR-Medicine: task and setup}
\label{app:nonverifiable}

In the main paper, we primarily focus on tasks with verifiable rewards. These tasks all have checkers comparing the response against a known answer or a set of test cases. In this section, we ask whether \methodname{} still helps when the reward is instead a score assigned by an LLM judge. We also test a second model family. We use RaR-Medicine~\citep{gunjal2025rubricsrewardsreinforcementlearning}, a set of open-ended medical questions where each question comes with its own rubric. A rubric is a list of criteria, each labelled Essential, Important, Optional, or Pitfall. To score a response, a judge model (\texttt{gpt-4o-mini}) marks each criterion as met or not met (for a Pitfall, as avoided or committed), and the reward is the weighted fraction of satisfied criteria, with weights $1.0$, $0.7$, $0.3$, and $0.9$ for the four categories. We use $6400$ training questions, $192$ questions as the GEPA validation set, and $256$ held-out test questions.

We train two base models, OLMo-3-7B-Instruct~\citep{olmo2026olmo3} and Qwen3-4B-Instruct-2507~\citep{qwen3}, both without thinking mode. RL follows the shared configuration of Appendix~\ref{app:hyperparams}: $32$ problems per step, $G=8$ rollouts per problem, learning rate $10^{-6}$, and maximum prompt and response lengths of $2048$ tokens each.
\methodname{} uses $K=4$ prompts, cycle length $T=6$, the seed prompt ``Answer the medical question.'', and \texttt{gpt-5.2} as the reflection LM. The dataset has no system message, so the GEPA prompt is prepended to the user message instead of being injected as a system message. The per-cycle GEPA budget is $96$ validation examples and $960$ metric calls for OLMo, and $192$ and $1920$ for Qwen. Every $20$ steps we evaluate on the $256$ test questions with greedy decoding, one response per question, and report the mean rubric reward. The \methodname{} arm is evaluated with its current best fast-weight prompt prepended, since that prompt is part of the policy it trains.

\subsection{Displacement versus reward}
\label{app:medicine-kl}

We measure $\mathrm{KL}(\pi_{\text{train}}\,\|\,\pi_{\text{base}})$ with the estimator of Section~\ref{sec:displacement}. For each checkpoint (saved every $50$ steps) we sample $16$ responses at temperature $1.0$ for each of $80$ test questions, score every token under both the trained and the base model, and average the per-token log-ratio. The \methodname{} checkpoints are evaluated \emph{without} any GEPA prompt, so both arms see byte-identical inputs and the number reflects only how far the slow weights have moved.

\begin{figure}[!htb]
    \centering
    \includegraphics[width=0.85\linewidth]{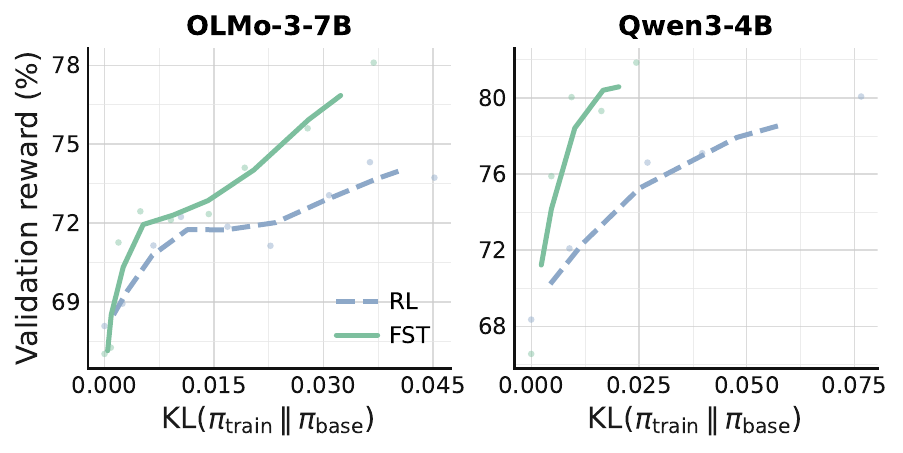}
    \caption{\textbf{Validation reward versus $\mathrm{KL}(\pi_{\text{train}}\,\|\,\pi_{\text{base}})$ on RaR-Medicine}, a task with an LLM-judged rubric reward rather than a verifier. \textbf{Left:} OLMo-3-7B-Instruct. \textbf{Right:} Qwen3-4B-Instruct-2507. Same conventions as Figure~\ref{fig:kl-vs-acc}: translucent markers are per-checkpoint measurements, the line is a rolling mean along training step, and each curve starts at the base model ($\mathrm{KL}=0$). Checkpoints every $50$ steps are shown, up to step $400$ for OLMo and step $200$ for Qwen. On both models \methodname{} (green) reaches higher reward at lower KL than RL (blue).}
    \label{fig:kl-vs-acc-medicine}
\end{figure}

Figure~\ref{fig:kl-vs-acc-medicine} shows the same pattern as the verifiable tasks. On Qwen3-4B, 
, the slow weights move about $3\times$ less for a higher reward. On OLMo-3-7B, after $500$ steps \methodname{} reaches $77.9\%$ versus $74.6\%$ for RL, at a KL of $0.059$ versus $0.071$. \methodname{} also needs fewer steps: it first matches the highest reward that RL attains during training after $120$ steps instead of $200$ on Qwen ($1.7\times$ fewer), and after $400$ steps instead of $480$ on OLMo ($1.2\times$ fewer). The gains here are smaller than on \texttt{HoVer} or \texttt{CodeIO}, which we attribute to the reward: a rubric score from a judge is a noisier and more graded training signal than a binary verifier, so the per-step feedback available to both the RL update and the GEPA reflection is weaker. With a judge-based reward and a second model family, \methodname{} still learns faster and stays closer to the base model than RL.

\subsection{Training curves}
\label{app:medicine-curves}

\begin{figure}[H]
    \centering
    \includegraphics[width=\linewidth]{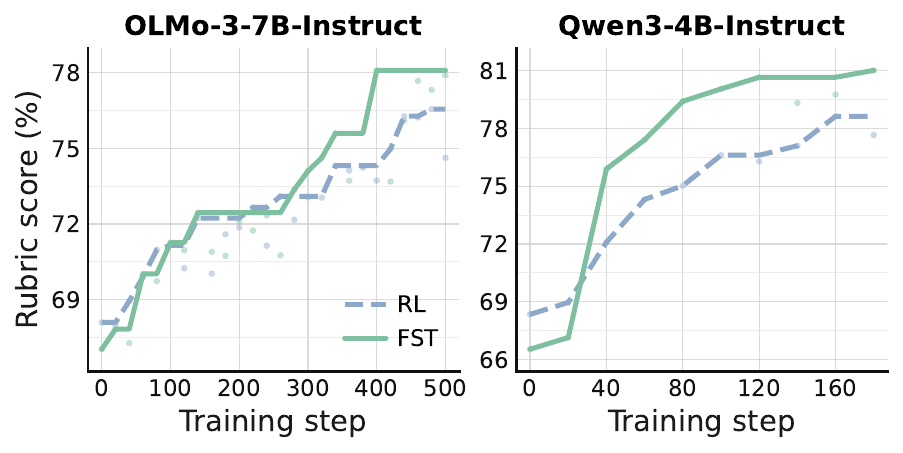}
    \caption{\textbf{Training curves for the RaR-Medicine runs}, as running
    maxima of the held-out rubric score; faint points are the raw
    per-evaluation scores. OLMo-3-7B-Instruct is shown for its full $500$ steps; Qwen3-4B-Instruct is shown to step $180$.}
    \label{fig:medicine-nonverifiable}
\end{figure}

Figure~\ref{fig:medicine-nonverifiable} shows the training curves for RaR-Medicine. On Qwen3-4B the two arms separate by step $40$ and the gap is
stable for the rest of training; on OLMo-3-7B they are indistinguishable for
roughly the first $300$ steps and separate only late, consistent with the
smaller step reduction.

\FloatBarrier
\section{Continual learning}
\label{app:sec:cl}

\subsection{Additional baselines}
\label{app:cl-baselines}

Section~\ref{sec:cl} compares \methodname{} with plain sequential RL. Here we add three RL variants that target the two usual explanations for a continual-learning gap, forgetting and parameter drift, plus one that gives RL the prompt channel, and we repeat the comparison under a second task ordering. The variants we run are: RL with experience replay, where $20\%$ of every batch is drawn from the training sets of earlier stages; RL with a KL penalty to the base model (coefficient $10^{-3}$) to limit displacement; and RL with one GEPA optimization at each task boundary, whose prompt is then frozen for the stage. Importantly, last variant is an unfair continual learning baseline and is kept for comparison only. The approach uses the task boundary as a signal for when to re-optimize the prompt. By contrast, \methodname{} is fully boundary agnostical and automatically changes it's context as the distribution of problems changes

\begin{figure}[!htb]
    \centering
    \includegraphics[width=\linewidth]{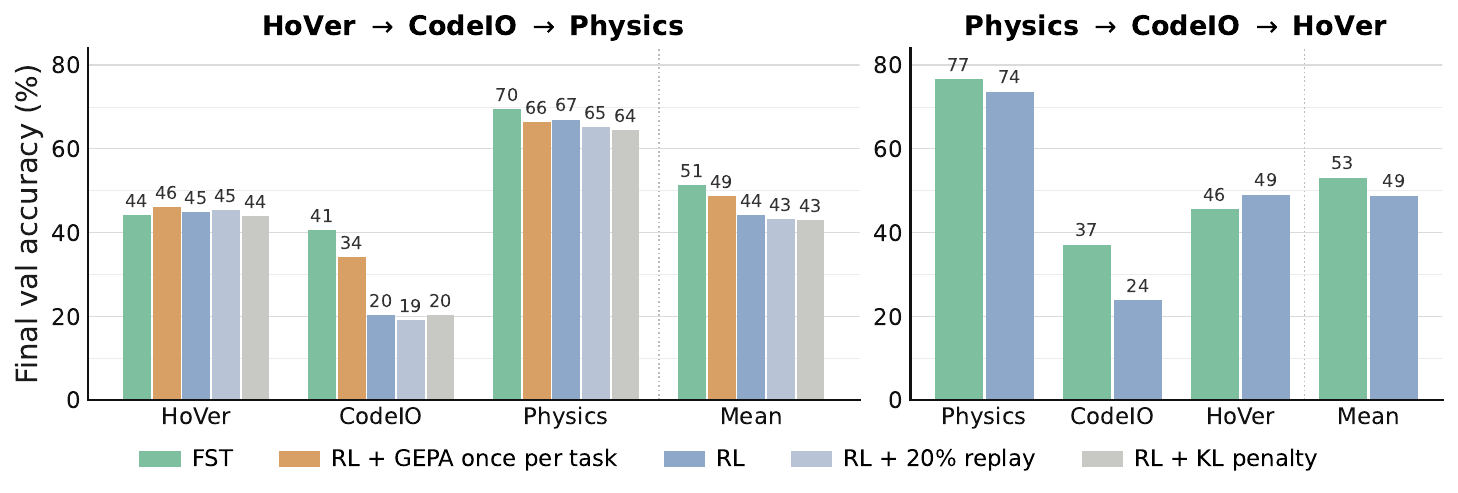}
    \caption{\textbf{Continual-learning baselines and a second ordering.} Final validation accuracy on each task at step $600$ of the three-stage run of Section~\ref{sec:cl} (one stage of $200$ steps per task, no optimizer reset), and the mean over the three tasks. (\textbf{Left}) HoVer $\to$ CodeIO $\to$ Physics: \methodname{} against plain RL and three RL variants. (\textbf{Right}) Physics $\to$ CodeIO $\to$ HoVer: plain RL against \methodname{}.}
    \label{fig:cl-baselines}
\end{figure}

Figure~\ref{fig:cl-baselines} (left) shows the results. Every arm retains HoVer at $44$--$46\%$, so forgetting does not separate the methods on this sequence. What separates them is the CodeIO stage, where plain RL stalls at $20.4\%$. Replay ($19.1\%$) and the KL penalty ($20.2\%$) do not help, so neither rehearsal nor a displacement constraint on its own explains \methodname{}'s gain. A prompt optimized once per task recovers most of the stall ($34.2\%$), and letting the prompt keep adapting during the stage recovers the rest ($40.6\%$ for \methodname{}). On the three-task mean, \methodname{} leads plain RL by $7.4$\,pp and the GEPA-once variant by $2.6$\,pp. Population size matters little here: \methodname{} with $K{=}2$, $4$, and $8$ reaches means of $49.0\%$, $51.4\%$, and $49.7\%$, all above every RL variant, so prompt diversity is not what drives the gap either.

\subsection{A second task ordering}
\label{app:cl-ordering}

Under the second ordering (Figure~\ref{fig:cl-baselines}, right), the CodeIO stall reproduces for plain RL ($23.8\%$) and \methodname{} again lifts it ($37.1\%$). \methodname{} also learns Physics slightly better ($76.6\%$ vs.\ $73.6\%$) while plain RL is ahead on the final HoVer stage ($49.0\%$ vs.\ $45.6\%$), for a three-task mean of $53.1\%$ against $48.8\%$.

\subsection{A longer stream of smaller tasks}
\label{app:rg-cl}

\begin{figure}[!htb]
    \centering
    \includegraphics[width=\linewidth]{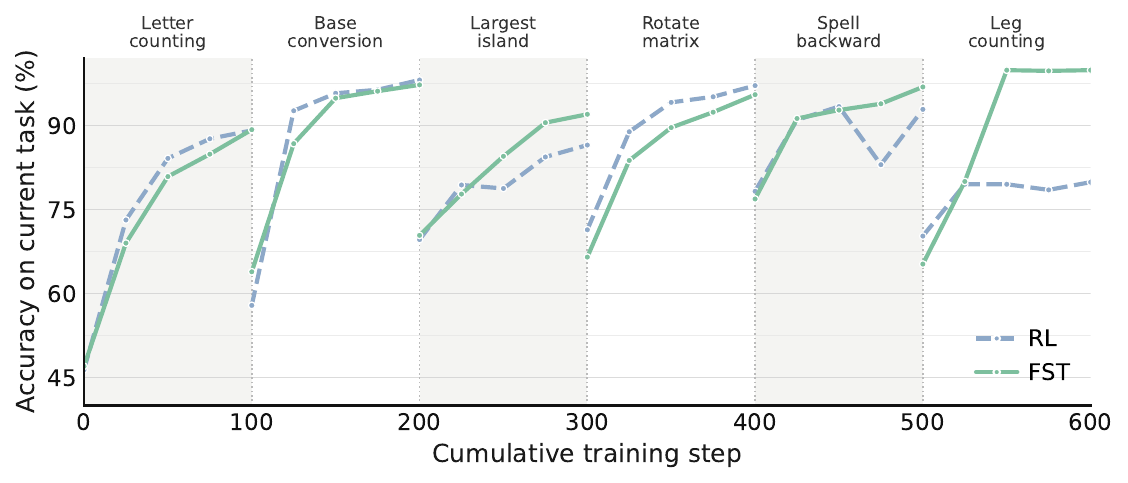}
    \caption{\textbf{A six-task reasoning-gym stream on Qwen3-4B.} Accuracy on the task currently being trained, evaluated every $25$ steps on its held-out set; the task switches every $100$ steps, so the curve restarts at each boundary. RL (dashed) and \methodname{} (solid) learn every task; \methodname{} ends three of the six stages higher and ties the rest.}
    \label{fig:rg-cl}
\end{figure}

The three-task sequences above use large, slow tasks. To see how the two methods behave over more task switches we also ran a stream of six reasoning-gym tasks~\citep{reasoning_gym} on Qwen3-4B (thinking mode): letter counting, base conversion, largest island, rotate matrix, spell backward, and leg counting, in that order, with $100$ RL steps per task and no optimizer reset at the switches. Each task has $3200$ training and $200$ held-out problems, the response budget is $4096$ tokens, and validation is run every $25$ steps on the task currently being trained. \methodname{} uses $K{=}4$, $T{=}6$, the user-prefix prompt position, and \texttt{gpt-5.2} as the reflection LM, as in the other continual-learning runs. Figure~\ref{fig:rg-cl} shows the two arms. Both learn every task within its $100$-step stage, and where they differ \methodname{} is ahead: it ends the stage higher on largest island ($+5.5$\,pp), spell backward ($+4.0$\,pp), and leg counting ($+20.0$\,pp, where RL plateaus at $80\%$ and \methodname{} reaches $99.9\%$), and is within $2$\,pp of RL on the other three tasks, for a mean end-of-stage gain of $+4.5$\,pp.

\FloatBarrier
\section{Experimental setup}
\label{app:sec:setup}

\subsection{GEPA} \label{appendix:gepa}
We optimize the fast weights $\phi$ using GEPA~\citep{gepa}, a reflective evolutionary procedure that searches the text space $\Sigma^\ast$ guided by a frozen \emph{reflection LM} $\pi_{\text{ref}}$, a separate capable model that proposes textual mutations from natural-language critiques of rollouts. For a candidate $\phi$ and query $x$, define the per-instance fitness \begin{equation} s(\phi; x) \;=\; \mathbb{E}_{y \sim \pi_\theta(\cdot \mid x, \phi)}\!\big[\,r(x, y)\,\big]. \label{eq:fitness-appendix} \end{equation} GEPA maintains a population $\mathcal{P}$ of candidate prompts whose fitness vectors $\big(s(\phi; x_1), \ldots, s(\phi; x_n)\big)$ on an anchor set $\{x_i\}_{i=1}^{n} \subset \mathcal{D}$ are tracked. One generation proceeds in four steps: (i) select a parent $\phi_p$ from the per-instance Pareto frontier of $\mathcal{P}$; (ii) sample a minibatch of rollouts under $\phi_p$ from $\pi_\theta$; (iii) elicit a child $\phi_c \leftarrow \pi_{\text{ref}}\!\big(\phi_p,\,\text{traces}\big)$ by having the reflection LM diagnose failures and propose a textual edit; (iv) evaluate $s(\phi_c; \cdot)$ on the anchor set, add $\phi_c$ to $\mathcal{P}$, and prune dominated candidates. After a fixed metric-call budget, GEPA returns the top-$m$ candidates of the resulting Pareto frontier. The frontier preserves diversity: different candidates are best on different slices of $\mathcal{D}$, and this diversity is precisely what allows the RL phase to exploit several complementary fast weights simultaneously. GEPA is related to other LLM-as-optimizer methods that use natural-language reflection or self-feedback~\citep{reflexion,selfrefine,opro}, and is distinguished by its per-instance Pareto population and explicit prompt-mutation operator.

\subsection{Algorithm pseudocode}
\label{app:algorithm}
We present the algorithm pseudocode in Algorithm~\ref{alg:method}.

\begin{algorithm}[h]
\caption{\methodname{}: interleaved RL with population-based prompt evolution.}
\label{alg:method}
\begin{algorithmic}[1]
\Require initial slow weights $\theta_0$; seed prompt $\phi_{\text{seed}}$; data stream $\mathcal{D}$; cycle length $T$; population size $K$; GRPO group size $G$ with $K \mid G$; reflection LM $\pi_{\text{ref}}$
\State $\Phi_0 \leftarrow \{\phi_{\text{seed}}\}$
\For{cycle $c = 0, 1, 2, \dots$}
    \State $\mathcal{L}_c \leftarrow$ pre-fetch the next $T$ minibatches from $\mathcal{D}$ \Comment{lookahead batch}
    \State $\Phi_{c+1} \leftarrow$ \textsc{Gepa}$\big(\pi_{\theta_c},\, \pi_{\text{ref}},\, \mathcal{L}_c,\, \Phi_c,\, K\big)$
    \For{$t = 1, \dots, T$}
        \State sample minibatch $\mathcal{B}_t \subset \mathcal{L}_c$
        \ForAll{$p \in \mathcal{B}_t$}
            \ForAll{$\phi^{(k)} \in \Phi_{c+1}$}
                \State assemble $G/K$ rollouts under $(p, \phi^{(k)})$, sampling from $\pi_\theta(\cdot \mid p, \phi^{(k)})$
            \EndFor
            \State place all $G$ rollouts in one group; compute group-relative advantages
        \EndFor
        \State update $\theta$ with $\mathcal{L}_{\textsc{cispo}}$ (Eq.~\eqref{eq:cispo})
    \EndFor
\EndFor
\end{algorithmic}
\end{algorithm}

\subsection{Star-graph dataset construction}
\label{app:star-graph}

Star-graph search is a planning task introduced in \citet{bachmann2025pitfallsnexttokenprediction} and used as a difficult search LLM RL problem in
\citet{prakash2025zerorewards}. We adopt their problem definition and procedurally
generate our train and test splits.

\paragraph{Graph instance.} Each instance is parameterized by a triple
$(d, p, n)$: source degree $d$, path length $p$, and node-pool size $n$. We
sample without replacement from $\{0, 1, \ldots, n-1\}$ to draw the source
$s$ and goal $g$ ($s \neq g$), then $p-2$ distinct intermediate nodes that
together form the unique gold path
$s \!\to\! v_1 \!\to\! \cdots \!\to\! v_{p-2} \!\to\! g$
(yielding $p-1$ edges of length 1 each). On top of the gold path we attach
$d-1$ \emph{decoy branches} rooted at $s$, each itself a chain of length
$p$, with all decoy nodes drawn fresh from the unused pool so that no decoy
intersects the gold path or another decoy. The full edge set is then
shuffled uniformly at random and serialized as a flat space-separated list
of comma-separated pairs. The graph is treated as undirected at scoring
time but presented as a list with no ordering hint.

\paragraph{Why this is a hard exploration problem.} The source $s$ is the
only node with degree $d$; every other node sits on a chain and has degree
$2$ (its predecessor and successor along that arm). The first hop is
therefore the only real branching decision: picking a decoy arm commits
the model to a chain that never reaches $g$, and the path-listing output
format gives no built-in way to backtrack. With $d=25$ a uniformly-random
first hop would succeed only $4\%$ of the time, and empirically
Qwen3-4B-Instruct does worse than uniform: pass@$16$ is $0/50$ on the seed
prompt before any RL update, because the model's strong path-finding prior
is miscalibrated for this synthetic layout and concentrates probability on
a wrong arm. This is the ``RL stuck at zero'' regime that motivates
fast-weight prompt evolution.

\paragraph{Prompt template.} Every example is formatted with the verbatim
template from the original reference implementation:
\begin{quote}\small\ttfamily
Given a bi-directional graph in the form of space separated edges,
output a path from source node to the destination node in the form of
comma separated integers.\\
For this question the graph is \{graph\}\\
The source node is \{source\}\\
The destination node is \{destination\}\\
Please reason step by step, and put your final answer within
\textbackslash boxed\{\}.
\end{quote}
The seed system prompt used during RL and as the initial GEPA candidate is:
\begin{quote}\small\itshape
You are solving a graph path-finding task. You will be given a list of
edges and a source and destination node. Output one valid path from source
to destination. Inspect the source node's neighbors first, identify which
neighbor leads to the destination via a sequence of valid edges, then
commit to that branch. Each consecutive pair in your output path must be
a valid edge in the graph. Put your final answer comma-separated inside
boxed braces.
\end{quote}

\paragraph{Scoring.} The reward function 
extracts the contents of the last \texttt{\textbackslash boxed\{\dots\}} from
the post-\texttt{</think>} body of the rollout, strips whitespace, and
applies an exact-match comparison against the gold path
$s, v_1, \ldots, v_{p-2}, g$ rendered as a comma-separated string. The
reward is $1.0$ on exact match and $0.0$ otherwise. The task admits no
partial credit, so even a single wrong intermediate node zeros the reward.

\paragraph{Splits used in the paper.} We sweep difficulty by varying
$(d, p, n)$. The headline experiments use $(d,p,n) = (25, 20, 500)$ with
$10{,}000$ training examples and $200$ held-out test examples.

\subsection{Hyperparameters and compute}
\label{app:hyperparams}

This section lists all training hyperparameters needed to reproduce the
numbers in the main paper. Settings shared across domains are described
first; per-domain overrides follow in
Tables~\ref{tab:hyper-rl}--\ref{tab:hyper-gepa}.

\paragraph{Shared RL configuration.} All RL runs are GRPO~\citep{deepseekmath}
with the \textsc{cispo} surrogate loss~\citep{minimaxm1,scalerl}
($\text{clip\_low}=1.0$, $\text{clip\_high}=3.0$), advantage normalization
by per-group standard deviation and no KL-to-reference penalty. The actor optimizer is AdamW (PyTorch defaults
$\beta_1=0.9$, $\beta_2=0.999$, weight decay $0$) at learning rate
$10^{-6}$ with a $10$-step linear warm-up.
Each RL step samples $G=8$ rollouts per problem with
\texttt{train\_batch\_size}$\,=32$ problems (so $256$ rollouts per step),
runs PPO updates with \texttt{ppo\_mini\_batch\_size}$\,=32$, and uses
tensor-parallel size $1$ for the rollout engine (vLLM). At evaluation
time we report mean@$4$ over four rollouts per validation prompt at
temperature $0.6$, top-$p$ $0.95$. All runs use Qwen3-8B \emph{thinking} mode
except star-graph (which uses the 4B Instruct model). RL-only arms use each task's fixed original prompt at both training and evaluation and never see a GEPA prompt; the arms that give RL an optimized prompt before or after training are reported separately in Appendix~\ref{app:static-gepa-rl}.

\paragraph{Shared GEPA configuration.} GEPA cycles use the same settings
for all domains: cycle length $T=6$ RL steps between GEPA optimizations
(equivalently, $\texttt{warmstart\_steps}=\texttt{rl\_steps\_per\_cycle}=6$);
all $K$ population prompts are scored on every question
(\texttt{prompts\_per\_question}$\,=K$) with advantage grouping
\emph{Option B} (\texttt{advantage\_grouping="question"}), so a problem's
$G$ rollouts are split $G/K$ per prompt and a single group statistic
$(\bar r_g, \sigma_g)$ is computed across all of them. The reflection LM
is OpenAI \texttt{gpt-5.2}, accessed through LiteLLM. Per-cycle GEPA
budgets are $\texttt{num\_eval\_examples}=192$ and
$\texttt{max\_metric\_calls}=960$ across all domains except polaris
($96$ and $1500$) and star-graph ($200$ and $960$). Candidate prompts are
injected as a system message; for datasets whose raw prompt already
contains a system message (HoVer, Physics, Math) we \emph{merge} the
GEPA prompt into the existing system role rather than stack a second one. Each GEPA
cycle outputs a Pareto-frontier population of size $K$ which seeds the
next cycle.

\paragraph{Per-domain overrides.} Tables~\ref{tab:hyper-rl}
and~\ref{tab:hyper-gepa} summarize the parameters that vary across
training domains.

\begin{table}[!htb]
\centering
\small
\setlength{\tabcolsep}{4pt}
\caption{Per-domain RL hyperparameters and base models.
$L_{\text{ctx}}$/$L_p$/$L_r$ are the maximum context, prompt, and response
lengths (tokens). Train batch is the number of problems per RL step;
rollouts/step is $\text{batch} \times G$.}
\label{tab:hyper-rl}
\begin{tabular}{@{}llcrrc@{}}
\toprule
Domain & Base model & $L_{\text{ctx}}$ / $L_p$ / $L_r$ & Batch & Rollouts/step & GPU util. \\
\midrule
HoVer-hard & Qwen3-8B (think) & 18944 / 4096 / 8192 & 32 & 256 & 0.6 \\
Physics    & Qwen3-8B (think) & 18944 / 4096 / 8192 & 32 & 256 & 0.7 \\
CodeIO     & Qwen3-8B (think) & 18944 / 4096 / 8192 & 32 & 256 & 0.6 \\
Math (Polaris) & Qwen3-8B-SFT$^\dagger$ (think) & 12288 / 4096 / 8192 & 64 & 512 & 0.7 \\
Star-graph & Qwen3-4B-Instruct (no think) & \phantom{0}8192 / 4096 / 4096 & 32 & 256 & 0.6 \\
\bottomrule
\end{tabular}\\[2pt]
{\footnotesize $^\dagger$ Polaris uses our own continued-SFT base (Qwen3-8B base further SFT'd on Nemotron
to recover math performance, since Qwen3-8B-Instruct is already saturated
on math). For Polaris only, GRPO advantages are not normalized by group
std 
and the train batch is
doubled to $64$ problems/step.}
\end{table}

\begin{table}[!htb]
\centering
\small
\caption{Per-domain GEPA hyperparameters. $K$ is the population size;
$G/K$ is the number of rollouts each candidate prompt produces per
problem within an RL group. Eval set / metric calls are the per-cycle
GEPA budgets.}
\label{tab:hyper-gepa}
\begin{tabular}{lcccrrl}
\toprule
Domain & $K$ & $G/K$ & Cycle $T$ & Eval set & Metric calls & Reflection LM \\
\midrule
HoVer-hard & 8 & 1 & 6 & 192 & \phantom{0}960 & \texttt{gpt-5.2} \\
Physics    & 4 & 2 & 6 & 192 & \phantom{0}960 & \texttt{gpt-5.2} \\
CodeIO     & 8 & 1 & 6 & 192 & \phantom{0}960 & \texttt{gpt-5.2} \\
Math (Polaris) & 4 & 2 & 6 & \phantom{0}96 & 1500 & \texttt{gpt-5.2} \\
Star-graph & 8 & 1 & 6 & 200 & \phantom{0}960 & \texttt{gpt-5.2} \\
\bottomrule
\end{tabular}
\end{table}

\paragraph{Compute and wall-clock.} All runs are submitted to a SLURM
cluster with $8 \times$ H100 (80GB) per node. The headline runs in the
main paper are single-node ($1\times8$ GPU); the Polaris $K=8$ ablation
is the only multi-node configuration ($4\times8 = 32$ GPUs). Mean
per-RL-step wall-clock under the headline configuration is $\sim 60$\,s
for RL-only and $\sim 100$\,s for \methodname{} without rollout reuse
(HoVer-hard, $K=8$). Enabling rollout reuse
(Section~\ref{app:rollout-reuse}) brings the RL-step cost down to
$\sim 47$\,s, slightly faster than RL-only at the step level, because
$\sim 1/3$ of the RL group's rollouts are served from the GEPA
evaluation cache rather than freshly generated. This figure covers the
RL training loop only. 

\end{document}